\documentclass[jair,twoside,11pt,theapa]{article}
\usepackage{jair, theapa, rawfonts}
\ShortHeadings{The Mathematics of Changing one's Mind}
{Jacobs}
\firstpageno{1}

\newif\ifignore 
\ignorefalse
\newcommand{\auxproof}[1]{
\ifignore\mbox{}\newline
\textbf{PROOF:} \dotfill\newline
{\it #1}\mbox{}\newline
\textbf{ENDPROOF}\dotfill
\fi}

\usepackage{amsmath}
\usepackage{amssymb}
\usepackage{amstext}
\usepackage{amsthm}
\usepackage{mathtools}
\usepackage{mathrsfs}
\usepackage{stmaryrd}
\usepackage{xspace}
\usepackage{enumerate}
\usepackage{url}
\usepackage{nicefrac}
\usepackage{fancybox}

\usepackage{xypic}
\usepackage[all,cmtip]{xy}
\xyoption{2cell}
\UseTwocells
\xyoption{tips}

\usepackage{tikz}
\usetikzlibrary{circuits.ee.IEC}

\newtheorem{defin}{Definition}[section] 
\newtheorem{theorem}[defin]{Theorem}
\newtheorem{lemma}[defin]{Lemma}
\newtheorem{proposition}[defin]{Proposition}

\theoremstyle{definition}
\newtheorem{definition}[defin]{Definition}
\newtheorem{remark}[defin]{Remark}

\newtheorem{example}[defin]{Example}

\newenvironment{myproof}[1][Proof]%
   { \begin{trivlist}%
     \item[\hskip \labelsep {\bfseries #1}.]%
   }%
   { \end{trivlist}%
   }

\renewcommand{\arraystretch}{1.3}
\newcommand{\QEDbox}{\square}
\newcommand{\QED}{\hspace*{\fill}$\QEDbox$}

\newcommand{\klafter}{\mathrel{\bullet}}

\DeclareSymbolFont{T1op}{T1}{cmr}{m}{n}
\SetSymbolFont{T1op}{bold}{T1}{cmr}{bx}{n}
\DeclareMathSymbol{\mathguilsinglleft}{\mathopen}{T1op}{'016}
\DeclareMathSymbol{\mathguilsinglright}{\mathclose}{T1op}{'017}

\newcommand{\idmap}[1][]{\ensuremath{\mathrm{id}_{#1}}}
\newcommand{\after}{\mathrel{\circ}}
\newcommand{\set}[2]{\{#1\;|\;#2\}}
\newcommand{\setin}[3]{\{#1\in#2\;|\;#3\}}

\newcommand{\ket}[1]{\ensuremath{|{\kern.1em}#1{\kern.1em}\rangle}}
\newcommand{\bigket}[1]{\ensuremath{\big|{\kern.1em}#1{\kern.1em}\big\rangle}}

\newcommand{\one}{\ensuremath{\mathbf{1}}}

\newcommand{\andthen}{\mathrel{\&}}
\newcommand{\distributionsymbol}{\mathcal{D}}
\newcommand{\Dst}{\distributionsymbol}
\newcommand{\Giry}{\mathcal{G}}
\newcommand{\indic}[1]{\mathbf{1}_{#1}}
\newcommand{\Prob}{\footnotesize \mathrm{Pr}}
\newcommand{\no}[1]{#1^{\scriptscriptstyle \bot}} 

\makeatletter
\newcommand{\mathoverlap}[2]{\mathpalette\mathoverlap@{{#1}{#2}}}
\newcommand{\mathoverlap@}[2]{\mathoverlap@@{#1}#2}
\newcommand{\mathoverlap@@}[3]{\ooalign{$\m@th#1#2$\crcr\hidewidth$\m@th#1#3$\hidewidth}}
\makeatother

\newsavebox\sbpto
\savebox\sbpto{\begin{tikzpicture}[baseline=-2.5pt]
            \filldraw[fill=white,draw=white] circle (1.4pt);
            \filldraw[fill=white,draw=black,line width=0.2pt] circle
(1.2pt);
                \end{tikzpicture}}
\newcommand\chanto{\mathrel{\ooalign{$\to$\cr
            \hfil\!$\usebox\sbpto$\hfil\cr}}}
            
\newcommand\kto[2]{#1 \chanto #2}

\begin{document}

\title{The Mathematics of Changing one's Mind, \\
   via Jeffrey's or via Pearl's update rule}



\author{\name Bart Jacobs \email bart@cs.ru.nl \\
       \addr Institute for Computing and Information Sciences,\\
       Radboud University, Nijmegen, The Netherlands
}

\maketitle

\begin{abstract} 
Evidence in probabilistic reasoning may be `hard' or `soft', that is,
it may be of yes/no form, or it may involve a strength of belief, in
the unit interval $[0,1]$. Reasoning with soft, $[0,1]$-valued
evidence is important in many situations but may lead to different,
confusing interpretations. This paper intends to bring more
mathematical and conceptual clarity to the field by shifting the
existing focus from specification of soft evidence to accomodation of
soft evidence.  There are two main approaches, known as Jeffrey's rule
and Pearl's method; they give different outcomes on soft evidence.
This paper argues that they can be understood as correction and as
improvement. It describes these two approaches as different ways of
updating with soft evidence, highlighting their differences,
similarities and applications. This account is based on a novel
channel-based approach to Bayesian probability. Proper understanding
of these two update mechanisms is highly relevant for inference,
decision tools and probabilistic programming languages.
\end{abstract}

\section{Introduction}\label{IntroSec}

Logical statements in a probabilistic setting are usually interpreted
as \emph{events}, that is, as subsets $E\subseteq \Omega$ of an
underlying sample space $\Omega$ of possible worlds, or equivalently
as characteristic functions $\Omega \rightarrow \{0,1\}$. One
typically computes the probability $\Prob(E)$ of an event $E$,
possibly in conditional form $\Prob(E\mid D)$ where $D$ is also an
event.  Events form the basic statements in probabilistic inference,
where they can be used as evidence or observation. Here we shall use a
more general interpretation of logical statements, namely as functions
$\Omega \rightarrow [0,1]$ to the unit interval $[0,1]$. They are
sometimes called fuzzy events or fuzzy predicates, but we simply call
them predicates.

The above description $\Omega \rightarrow \{0,1\}$ of events/evidence
is standard. It is sometimes called hard or certain or sharp evidence,
in contrast to soft, uncertain, unsharp, or fuzzy evidence $\Omega
\rightarrow [0,1]$. In most textbooks, see
\textit{e.g.}~\cite{Barber12,BernardoS00,JensenN07,KollerF09,Pearl88}
on Bayesian probability, soft evidence is missing or is only a
marginal topic.  For instance, in~\cite[\S3.2]{Barber12} it is
discussed only briefly, namely as: ``In soft or uncertain evidence,
the evidence variable is in more than one state, with the strength of
our belief about each state being given by probabilities.'' The topic
gets relatively much attention in~\cite[\S3.6-3.7]{Darwiche09},
starting from a description: ``Hard evidence is information to the
effect that some event has occurred ... Soft evidence, on the other
hand, is not conclusive: we may get an unreliable testimony that event
$\beta$ occurred, which may increase our belief in $\beta$ but not to
the point where we would consider it certain.''


Typically, soft evidence deals with statements like: I saw the object
in the dark and I am only $70\%$ sure that its color is red. Or: my
elder neighboor has hearing problems and is only $60\%$ certain that
my alarm rang. As said, we interpret such evidence as fuzzy
predicates, with a degree of truth in $[0,1]$.  Somewhat confusingly,
these statements may also be interpreted as a state of affairs, that
is as a probability distribution with a convex combination of $0.7$
red and $0.3$ non-red. It seems fair to say that there is no widely
accepted perspective on how to interpret and reason with such soft
evidence, and in particular on how to update with soft evidence. The
mathematics of such updating is the main topic of this paper, which,
in the words of~\cite{DiaconisZ83}, is called: the mathematics of
changing one's mind.

In fact, there are two main approaches to soft updating, that is, to
updating with soft evidence. They are most clearly distinguished
in~\cite{ChanD05}, but see
also~\cite{Darwiche09,DiaconisZ82,ValtortaKV02}.
\begin{enumerate}
\item One can use \emph{Jeffrey's rule}, from~\cite{Jeffrey83}, see
  also~\cite{Halpern03,Shafer81}. It interprets softness as a
  probability distribution that represents a new state of affairs that
  differs from what is predicted, and that one needs to \emph{adjust}
  or \emph{correct} to. Adjusting to $70\%$ probability of seeing red
  involves a convex combination of point updates: one takes $0.7$
  times the belief revision for red plus $0.3$ times the revision for
  not-red. This approach focuses on adjustment/correction to a new
  state of affairs. Phrases associated with this approach are
  `probability kinematics'~\cite{Jeffrey83}, `radical
  probabilism'~\cite{Skyrms96}, or dealing with
  `surprises'~\cite{DietrichLB16} or with `unanticipated
  knowledge'~\cite{DiaconisZ83}.


\item One can also use \emph{Pearl's method of virtual evidence},
  from~\cite{Pearl88,Pearl90}. This approach is described
  operationally: extend a Bayesian network with an auxiliary node, so
  that soft evidence can be emulated in terms of hard evidence on this
  additional node, and so that the usual inference methods can be
  applied.  We shall see that extending a Bayesian network with such a
  node corresponds to using a fuzzy predicate to capture the soft
  evidence. This approach \emph{factors in} the soft evidence,
  following the basic idea: $\mathit{posterior} \propto \mathit{prior}
  \cdot \mathit{likelihood}$. It involves improvement instead of
  correction.
\end{enumerate}

This paper takes a fresh mathematical perspective on a problem that
exists already for a long time in probabilistic reasoning, going back
to~\cite{Jeffrey83,Pearl90}. This work builds on a novel approach to
Bayesian probability theory, based on programming language semantics
and ultimately on category theory, see~\cite{Giry82} for an early
source, and~\cite{Jacobs18c} for a modern overview. This approach
clearly separates (fuzzy) predicates (evidence) from probability
distributions (states). It is therefor well-prepared to deal with
softness/uncertainty, either as fuzzy predicate or as state of
affairs. In our general reformulation, Pearl's rule uses a
\emph{predicate} as soft evidence and involves backward inference via
predicate transformation (see Definition~\ref{def:constructive}), in a
known manner, see~\cite{JacobsZ16,JacobsZ19}. The main (novel)
mathematical observation of this paper is that Jeffrey's rule is
captured via a \emph{state} (distribution) as soft evidence and via
state tranformation with the Bayesian inversion
(`dagger')~\cite{ClercDDG17} of the channel at hand, see
Definition~\ref{def:destructive}.

One fundamental problem is that the terminology in this area is
confusing and is not used consistently by various
authors. Reference~\cite{MradDPLA15} gives a good overview of the
different terminologies and their meaning (and of the literature on
this topic). It uses the terminology `likelihood evidence' or
`uncertain evidence' as `evidence \emph{with} certainty' for what we
call a predicate; it also uses `soft evidence' as `evidence \emph{of}
uncertainty' for a probability distribution. We shall build on the
distinction between predicates and states, since both notions are
mathematically well-defined (see below); we shall use evidence and
probability distribution as alternative names for predicate and
state. The adjectives soft, uncertain, fuzzy will be used here only in
an informal sense, without making a distinction between them. This
leads to the following table.
\begin{center}
\begin{tabular}{c || c | c}
\setlength{\tabcolsep}{5em}
\mbox{\quad Here \quad}
&
\begin{tabular}{c}
predicate \\[-0.4em]
evidence
\end{tabular}
&
\begin{tabular}{c}
state \\[-0.4em]
probability distribution 
\end{tabular}
\\
\hline
In~\cite{MradDPLA15}
&
\hspace*{0.5em}\begin{tabular}{c}
likelihood evidence \\[-0.4em]
uncertain evidence \\[-0.4em]
evidence with uncertainty
\end{tabular}\hspace*{0.5em}
&
\hspace*{0.5em}\begin{tabular}{c}
soft evidence \\[-0.4em]
evidence of uncertainty
\end{tabular}\hspace*{0.5em}
\end{tabular}
\end{center}

\noindent In accordance with this table, we shall say that Pearl's
update rule is evidence-based and Jeffrey's rule is state-based.

The literature on soft updating, see esp.~\cite{ChanD05,Darwiche09}
(and references given there), focuses on the way in which softness is
specified.  Quoting from~\cite{ChanD05}: ``The difference between
Jeffrey's rule and Pearl's method is in the way uncertain evidence is
specified. Jeffrey requires uncertain evidence to be specified in
terms of the \emph{effect} it has on beliefs once accepted, which is a
function of both evidence strength and beliefs held before the
evidence is obtained. Pearl, on the other hand, requires uncertain
evidence to be specified in terms of its \emph{strength} only.''  This
paper shifts the emphasis from \emph{specification} of softness to
\emph{accomodation} of softness, that is, to the precise update rules,
see Definition~\ref{def:constructive} and~\ref{def:destructive}, using
both predicates and states to capture softness. In the end, after
Lemma~\ref{lem:deterministic}, we demonstrate that specification in
terms of the update effect only works in the deterministic case. It is
thus not a method that can be used in general.

Some more technical background: within the compositional programming
language perspective, a Bayesian network is a (directed acyclic) graph
in the Kleisli category of the distribution monad $\Dst$ --- or the
Giry monad $\Giry$ for continuous probability theory ---
see~\cite{Fong12}.  The maps in these Kleisli categories are also
called \emph{channels}; they carry lots of useful algebraic structure
that forms the basis for a compositional approach to probability.
Along these channels one can do state transformation and predicate
transformation, like in programming language semantics. These
transformations are of direct relevance in Bayesian
inference~\cite{JacobsZ16,JacobsZ19}, giving rise, for instance, to a
new inference algorithm~\cite{Jacobs18b}. This paper builds on this
`channel-based' approach to give a novel precise account of Jeffrey's
and Pearl's update rules. However, no familiarity with category theory
is assumed and all the relevant concepts are introduced here.

The paper starts by elaborating a standard Bayesian example of a
disease, with a prior probability, and a test for the disease that has
a certain sensitivity; the question is: what can we infer about the
disease if we are $80\%$ sure the test comes out positive? We
illustrate how to compute the different outcomes of Jeffrey and Pearl
(12\% versus 3\% disease probability). We postpone the mathematical
analysis and first go into reflective mode in
Section~\ref{sec:observations}. There we consider the question how to
understand and when to use Jeffrey's or Pearl's approach. This leads
to a terminological table~\eqref{table:naming}. The mathematics itself
is precise and clear, see from Section~\ref{sec:channel} onwards, but
it often remains unclear when to use which approach. Our terminology
of `adjusting to a new state of affairs' (Jeffrey) versus `factoring
in new evidence' (Pearl) is meant to provide some guidance, but we are
the first to admit that this remains vague --- see also
Example~\ref{ex:Dietrich} copied from~\cite{DietrichLB16}, where both
approaches are used, for different
reasons. Section~\ref{sec:observations} briefly mentions some further
perspectives. For instance, if you perform Jeffrey's updating of your
belief with what you can predict you learn nothing new; if you do
Pearl's updating with no information (a uniform likelihood), you learn
nothing new. Both make sense, but they are clearly different. This
reflective section is meant chiefly to generate further discussion on
this fundamental and intriguing topic, but not to provide a decision
mechanism for the `right' form of updating.

In Section~\ref{sec:channel} the mathematical analysis starts. First,
background information is given about states, predicates, updating,
channels, and transformation along channels. This allows us to
identify Pearl's rule as backward
inference. Section~\ref{sec:inversion} first explains the Bayesian
inversion of a channel and then uses this construction to capture
Jeffrey's rule. Subsequently, Section~\ref{sec:literature} reviews
some standard examples from the literature in terms of the new
channel-based framework, and then shows how the earlier methods
focused on specification of soft evidence --- also known as ``all
things considered'' and ``nothing else considered''
after~\cite{GoldszmidtP96} --- fit naturally in the new framework.

\section{A simple illustration}\label{sec:illustration}

\begin{figure}
\[ \vcenter{\xymatrix@C-1pc@R-1pc{
{\setlength\tabcolsep{0.2em}
   \renewcommand{\arraystretch}{1}
\begin{tabular}{|c|}
\hline
disease occurrence \\
\hline\hline
$\nicefrac{1}{100}$ \\
\hline
\end{tabular}}
& \ovalbox{\strut disease}\ar[d] 
\\
{\setlength\tabcolsep{0.2em}
   \renewcommand{\arraystretch}{1}
\begin{tabular}{|c|c|}
\hline
disease & positive test \\
\hline\hline
$d$ & $\nicefrac{9}{10}$ \\
\hline
$\no{d}$ & $\nicefrac{1}{20}$ \\
\hline
\end{tabular}}
& \ovalbox{\strut test}
}}
\hspace*{5em}
\vcenter{\xymatrix@C-1pc@R-1pc{
\ovalbox{\strut disease}\ar[d] 
\\
\ovalbox{\strut test}\ar[d] 
\\
\ovalbox{\strut certainty}
  \rlap{\hspace*{2em}\smash{\setlength\tabcolsep{0.2em}
\renewcommand{\arraystretch}{1}
\begin{tabular}{|c|c|}
\hline
test & certainty \\
\hline\hline
$t$ & $r$ \\
\hline
$\no{t}$ & $1-r$ \\
\hline
\end{tabular}}}
}}
\hspace*{8em} \]
\caption{A Bayesian network for testing for a disease, on the left,
  and an extension of this network with a certainty node, where the
  relevant probability is a parameter $r\in[0,1]$.}
\label{fig:diseaseBN}
\end{figure}
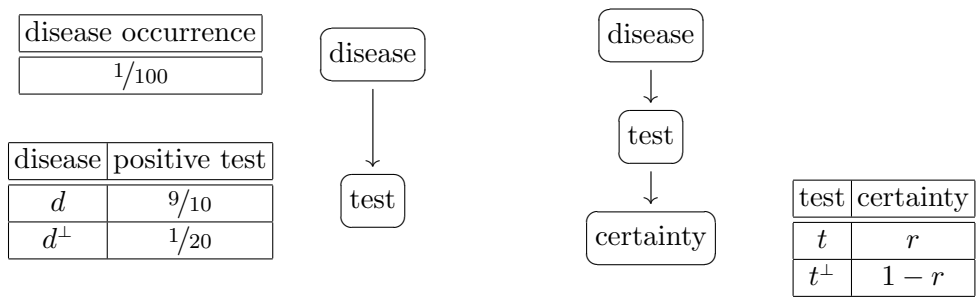

Consider a simple Bayesian network involving a test for a disease, as
on the left in Figure~\ref{fig:diseaseBN}. There is an a priori
disease probability of $1\%$. The test has a sensitivity as given by
the table on the lower-left in the figure: in presence of the disease,
written as $d$, the likelihood of a positive test outcome is $90\%$;
in absence of the disease, there is still a chance of $5\%$ that the
test comes out positive.

In this situation we can compute the predicted positive-test
likelihood $\Prob(t)$ via the law of total probability, as:
\[ \begin{array}{rcl}
\Prob(t)
& = &
\Prob(t \mid d)\cdot \Prob(d) + \Prob(t \mid \no{d})\cdot \Prob(\no{d})
\\
& = &
\frac{9}{10}\cdot\frac{1}{100} + \frac{1}{20}\cdot\frac{99}{100}
\hspace*{\arraycolsep} = \hspace*{\arraycolsep}
\frac{117}{2000}
\hspace*{\arraycolsep} \sim \hspace*{\arraycolsep}
6\%.
\end{array} \]

\noindent The probability of the disease $d$, given a positive test
$t$, is computed via Bayes' rule:
\[ \begin{array}{rcccccccl}
\Prob(d\mid t)
& = &
\displaystyle\frac{\Prob(t \mid d)\cdot \Prob(d)}{\Prob(t)}
& = &
\displaystyle\frac{\nicefrac{9}{10}\cdot\nicefrac{1}{100}}{\nicefrac{117}{2000}}
& = &
\displaystyle\frac{18}{117}
& \sim &
15\%
\end{array} \]

\noindent Similarly one obtains the conditional probability
$\Prob(d\mid \no{t}) =
\frac{\nicefrac{1}{1000}}{\nicefrac{1883}{2000}} = \frac{2}{1883} \sim
0.1\%$ of the disease given a negative test $\no{t}$.

\auxproof{
\[ \begin{array}{rcl}
\Prob(d\mid \no{t})
& = &
\displaystyle\frac{\Prob(\no{t} \mid d)\cdot \Prob(d)}{\Prob(\no{t})}
\\
& = &
\displaystyle\frac{\Prob(\no{t} \mid d)\cdot \Prob(d)}
   {\Prob(\no{t} \mid d)\cdot \Prob(d) + 
   \Prob(\no{t} \mid \no{d})\cdot \Prob(\no{d})}
\\[+1em]
& = &
\displaystyle\frac{\nicefrac{1}{10}\cdot\nicefrac{1}{100}}
   {\nicefrac{1}{10}\cdot\nicefrac{1}{100} + 
   \nicefrac{19}{20}\cdot\nicefrac{99}{100}}
\hspace*{\arraycolsep}=\hspace*{\arraycolsep}
\frac{\nicefrac{1}{1000}}{\nicefrac{1883}{2000}}
\hspace*{\arraycolsep}=\hspace*{\arraycolsep}
\frac{2}{1883}
\end{array} \]
}

This paper focuses on \emph{soft} evidence. It arises for instance in
a situation where the test outcome is observed in the dark, and that
there is, say, only $80\%$ certainty when the test is positive (and
$20\%$ certainty when it is negative).

There are two ways in the literature for handling soft evidence,
called Jeffrey's rule and Pearl's method for virtual
evidence. Jeffrey's rule says that we should take the
convex combination, with factors $0.8$ and $0.2 = 1 - 0.8$, of the
``point updates'', for the point evidence $t$ and $\no{t}$. Thus one
takes the convex combination of the above outcomes $\Prob(d\mid t)$
and $\Prob(d\mid \no{t})$, resulting in the probability:
\begin{equation}
\label{eqn:disease:DU}
\begin{array}{rcl}
0.8\cdot \Prob(d\mid t)  + 0.2\cdot \Prob(d\mid \no{t})
& \sim &
12\%.
\end{array}
\end{equation}

\noindent Thus, the certainties --- $80\%$ for a positive test and
thus $20\%$ for a negative test --- are used as weights for the two
corresponding conditional probabilities $\Prob(d\mid t)$ and
$\Prob(d\mid \no{t})$.  This makes sense.

In contrast, Pearl's rule involves extending the Bayesian network with
an additional binary node for `certainty', as on the right in
Figure~\ref{fig:diseaseBN}.  One can then compute the probability of
the disease if the test is positive with $80\%$ certainty in the usual
Bayesian way --- by taking $r = 0.8$ in the lower-right table in
Figure~\ref{fig:diseaseBN}:
\begin{equation}
\label{eqn:disease:CU}
\begin{array}{rcccccl}
\Prob(d\mid c)
& = &
\Prob(d\mid t)\cdot \Prob(t\mid c) + 
   \Prob(d\mid \no{t})\cdot \Prob(\no{t}\mid c) 
& = &
\frac{148}{4702}
& \sim &
3\%.
\end{array}
\end{equation}

\auxproof{
\[ \begin{array}{rcl}
\Prob(d\mid c)
& = &
\Prob(d\mid t)\cdot \Prob(t\mid c) + 
   \Prob(d\mid \no{t})\cdot \Prob(\no{t}\mid c) 
\\[+0.2em]
& = &
\displaystyle\frac{18}{117}\cdot 
   \frac{\Prob(c\mid t)\cdot \Prob(t)}{\Prob(c)} +
   \frac{2}{1883} \cdot 
   \frac{\Prob(c\mid \no{t})\cdot \Prob(\no{t})}{\Prob(c)}
\\[+1em]
& = &
\displaystyle\frac{18}{117}\cdot 
   \frac{\nicefrac{8}{10}\cdot \nicefrac{117}{2000}}
   {\nicefrac{8}{10}\cdot \nicefrac{117}{2000} + 
    \nicefrac{2}{10}\cdot \nicefrac{1883}{2000}} +
   \frac{2}{1883} \cdot 
   \frac{\nicefrac{2}{10}\cdot \nicefrac{1883}{2000}}
   {\nicefrac{8}{10}\cdot \nicefrac{117}{2000} + 
    \nicefrac{2}{10}\cdot \nicefrac{1883}{2000}}
\\[+1em]
& = &
\displaystyle\frac{18}{117}\cdot \frac{8 \cdot 117}{8 \cdot 117 + 2 \cdot 1883}
   + \frac{2}{1883} \cdot \frac{2 \cdot 1883}{8 \cdot 117 + 2 \cdot 1883}
\\[+1em]
& = &
\displaystyle\frac{8\cdot 18}{4702} + \frac{4}{4702} 
\\[+1em]
& = &
\displaystyle\frac{148}{4702}
\hspace*{\arraycolsep}\sim\hspace*{\arraycolsep}
3 \%
\end{array} \]
}

\noindent This approach also makes sense. But its outcome differs
substantially from Jeffrey's outcome of $12\%$
in~\eqref{eqn:disease:DU}. Which rule is the right one here: Jeffrey's
or Pearl's?

\begin{figure}
\begin{center}
\includegraphics[scale=0.36]{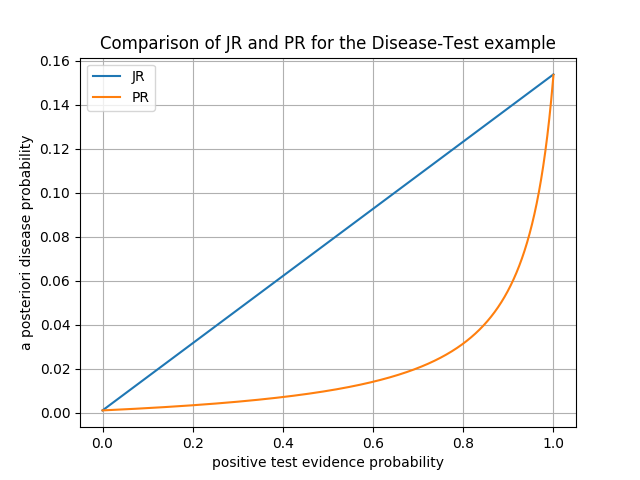}
\quad
\includegraphics[scale=0.36]{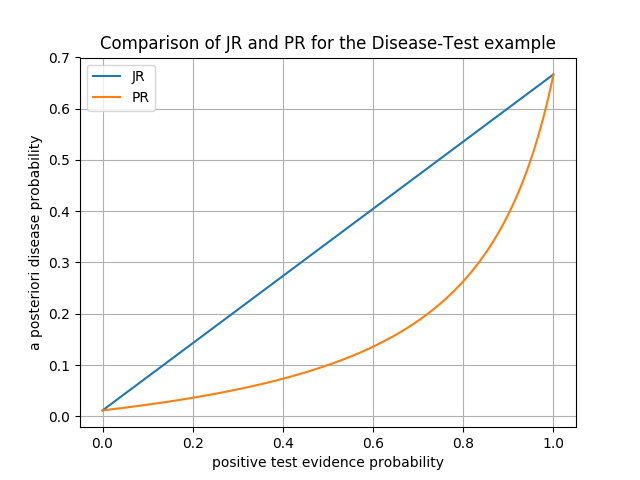}
\end{center}
\caption{Probabilities obtained by applying both Jeffrey's rule (JR)
  and Pearl's rule (PR) to the disease-test network of
  Figure~\ref{fig:diseaseBN}. The plot on the left captures the
  outcomes for an a priori disease probability of $1\%$, whereas the
  plot on the right uses $10\%$ instead.}
\label{fig:diseasePlots}
\end{figure}

In order to get a better picture we now take the soft evidence
probability for a positive test as a parameter $r\in [0,1]$. Thus the
number $r$ represents the certainty of a positive test. The resulting
a posteriori disease probabilities are plotted in
Figure~\ref{fig:diseasePlots}, on the left for the a priori disease
probability of $1\%$, and on the right for the higher prior of $10\%$.
We see that the two lines coincide at the extremes, for $r=0$ and
$r=1$, corresponding to $100\%$ certainty of a negative test and
$100\%$ certainty of a positive test. Inbetween these extremes, when
$0 < r < 1$, the outcomes differ. Thus, making a distinction between
the use of a Jeffrey's and Pearl's rule really only makes sense for
soft evidence.

We also see that Jeffrey's rule yields a straight line. This is
because it is defined by the linear (convex) function:
\[ \begin{array}{rcl}
r
& \longmapsto &
r\cdot \Prob(d\mid t)  + (1-r)\cdot \Prob(d\mid \no{t}).
\end{array} \]

\noindent Pearl's rule gives a non-linear outcome, according to the
familiar formula: $\mathit{posterior} \propto \mathit{prior} \cdot
\mathit{likelihood}$.

A final observation is that both rules do take the prior into account:
the range of outcomes is quite different on the left and on the right.

\section{Some observations about Jeffrey's and Pearl's 
   rules}\label{sec:observations}

Before focusing on a mathematical analysis we like to make some
remarks about the delicate question which form of updating ---
Jeffrey's or Pearl's --- is the `right' one.  It is an important
issue, for instance in the implementation of inference tools
(see~\cite{MradDPLA15} for an overview) or decision support systems,
since as we have seen in the previous section, the two approaches give
radically different outcomes.

The question `which rule is the right one' may be refined to: under
which circumstances should we use which rule, with which interpretation
of softness?

Here we propose the following intuitive explanation of the two
approaches, applied to the disease-test example from the previous
section, where, recall, we had $80\%$ certainty about a postive test
outcome.
\begin{itemize}
\item Jeffrey's approach is state-based and uses the $80\%$
  positive-test certainty as a probability distribution (state), for
  which we shall use the following notation: $0.8\ket{t} +
  0.2\ket{\no{t}}$. This means that we interpret it as a given
  \emph{state of affairs} in which the test has a positive outcome $t$
  with a likelihood of $80\%$ and a negative test outcome $\no{t}$
  with $20\%$ probability. When we see this state of affairs as a new
  situation --- a `surprise' as suggested in~\cite{DietrichLB16} ---
  and we wish to \emph{adjust} or \emph{correct} to this state of
  affairs, we use Jeffrey's rule as a form of backtracking.


\item Pearl's approach is evidence-based: the $80\%$ certainty is used
  as uncertain evidence that is \emph{factored in}, via a suitable
  multiplication with the prior information (plus normalisation).  The
  evidence is not treated as surprising, but as additional information
  that is smoothly taken into account, in the regular Bayesian
  manner. This can be described either via an extra variable, or via a
  predicate, see Section~\ref{sec:channel}.
\end{itemize}

\noindent The suggestion here is that Jeffrey's rule is for \emph{correction}
and Pearl's rule for \emph{improvement}. The following table 
summarises the terminology.
\begin{equation}
\label{table:naming}
\begin{tabular}{c || c | c}
\setlength{\tabcolsep}{5em}
\textbf{\qquad rule\qquad} & \textbf{\qquad uncertainty via\qquad} & 
  \textbf{\quad form of updating\quad}
\\
\hline\hline
\begin{tabular}{c}
Jeffrey's 
\end{tabular}
&
\begin{tabular}{c}
state of affairs \\[-0.4em]
proability distribution
\end{tabular}
&
\begin{tabular}{c}
state-based \\[-0.4em]
adjusting to \\[-0.4em]
correction 
\end{tabular}
\\
\hline
\begin{tabular}{c}
Pearl's 
\end{tabular}
&
\begin{tabular}{c}
predicate \\[-0.4em]
evidence 
\end{tabular}
&
\begin{tabular}{c}
evidence-based \\[-0.4em]
factoring in \\[-0.4em]
improvement 
\end{tabular}
\end{tabular}
\end{equation}

\noindent The remainder of this section contains some general
observations and questions for further research.
\begin{enumerate}
\item From a mathematical perspective, Pearl's update rule is most
  well-behaved. In particular, iterated applications of the
  constructive rule commute, see
  Proposition~\ref{prop:constructive}~\eqref{prop:constructive:and},
  whereas multiple usages of Jeffrey's rule do not commute. This is in
  line with the idea that the Jeffrey's approach involves abrupt
  adjustments.

\smallskip

\item Pearl's rule makes classical use of Bayesian networks, as is
  illustrated via the additional binary node in
  Figure~\ref{fig:diseaseBN}, on the right. In inference in such
  networks one factors in the evidence by propagating it through the
  network --- and then marginalising.

\smallskip

\item In certain (other) cases one may explicitly wish to have an
  alternative rule for updating. For instance,~\cite{ValtortaKV02}
  describes a model of multi-agent systems, each with their own
  knowledge represented via a local Bayesian network. It is explicitly
  required that: ``The mechanism for integrating the view of the other
  agents on a shared variable is to replace the agent's current belief
  in this variable with that of the communicating agent.'' Such
  replacements are obtained via Jeffrey's rule.

\smallskip

\item One can try to think of experimental verifications of the rules
  of Jeffrey/Pearl. A frequentist approach involves computing ratio's
  via counting and seems to support Jeffrey's form of updating. After
  all, Jeffrey's rule involves taking a convex sum of updates with
  individual point observations.

\smallskip

\item If probabilistic updating is seen as a mathematical model (or
  approximation) of cognitive priming, see
  \textit{e.g.}~\cite{GriffithsKT08}, then the non-commutativity of
  iterated applications of Jeffrey's rule may be seen as a good
  thing. Indeed, the human mind is sensitive to the order in which it
  receives evidence, that is, in which it is being primed. This `order
  effect' of priming can be illustrated in simple examples. The
  author's favourite one is: what image arises in your mind from the
  following two sequences of sentences?
\begin{center}
Alice is pregnant; Bob visits Alice
\\
\textit{versus}
\\
Bob visits Alice; Alice is pregnant.
\end{center}

\noindent Maybe cognitive psychologists can provide more clarity about
wether Jeffrey's or Pearl's rule works best in their field, see also
the suggested connection to~\cite{Hohwy13} in
Section~\ref{sec:conclusion}.

\smallskip

\item If probabilistic updating is seen as a form of learning --- in
  an informal sense, not as parameter/structure learning --- then one
  can ask what is the best model for handling evidence: correction of
  existing knowledge, as in the Jeffrey's approach, or improvement of
  existing knowledge, as in Pearl's approach. The question also comes
  up by comparing
  Propositions~\ref{prop:constructive}~\eqref{prop:constructive:effect}
  and~\ref{prop:destructive}~\eqref{prop:destructive:effect}.  They
  can be read informally as follows.

\smallskip

\begin{enumerate}
\item Pearl's improvement-based rule says: if you update (improve)
  your belief with no information (a uniform likelihood), you learn
  nothing new.

\item Jeffrey's correction-based rule says: when you update (correct)
  your belief with what you already know, you learn nothing new.
\end{enumerate}

\noindent Both readings make sense and connect the informal reading
(improvenment versus correction) to mathematical facts.

\smallskip

\item One might think that the distinction Jeffrey/Pearl is related to
  whether or not the base rate (prior distribution) is taken into
  account in probabilistic reasoning. As shown in~\cite{TverskyK82},
  people are not very good at doing so. But the outcomes of both rules
  do depend on the prior, see the (vertical scales of the) two plots
  in Figure~\ref{fig:diseasePlots}.

\smallskip

\item In the end, one can imagine using a combination of Jeffrey's
  rule (JR) and a Pearl's rule (PR), via a convex sum
\[ s\cdot\text{JR} + (1-s)\cdot\text{PR} \]

\noindent The number $s\in [0,1]$ then captures the novelty of the
evidence. Very speculatively, it may be related to the degree to which
the evidence's effect is absorbed (in one's brain).
\end{enumerate}

\section{Channel-based probabilistic reasoning}\label{sec:channel}

This section lays the foundation for our mathematical description of
soft updating, using either Pearl's or Jeffrey's rule. Traditionally
in probabilistic logic \emph{events} are used as evidence. Such events
form subsets $E\subseteq X$ of the sample space $X$; they correspond
to characteristic functions $\indic{E} \colon X \rightarrow \{0,1\}$,
defined by $\indic{E}(x) = 1$ iff $x\in E$.  As is well-known, these
events (subsets) form a Boolean algebra. In order to deal with
softness we use more general `fuzzy' predicates, of the form $p\colon
X \rightarrow [0,1]$, sending each element $x\in X$ to a probability
$p(x)\in [0,1]$, representing the strength of belief. Such a predicate
is called `likelihood evidence' in~\cite{MradDPLA15} or a `fuzzy
event' in~\cite{Zadeh68} where $p(x)$ described the `grade of
membership'. These predicates do not form a Boolean algebra, but what
is called an effect module, see
\textit{e.g.}~\cite{Jacobs15d,Jacobs18c}.

Below we sketch a reformulation of the basics of probabilistic
reasoning, in order to systematically accomodate soft/uncertain/fuzzy
evidence. This reformulation uses basic mathematical concepts like
distribution (state), fuzzy predicate, channel, conditioning, state-
and predicate-transformation. These concepts stem from the area of
program semantics where a distinction between predicate-transformer
and state-transformer semantics is common, see
\textit{e.g.}~\cite{DijkstraS90,Kozen85}. For a more extensive
introduction of these concepts in probabilistic reasoning we refer
to~\cite{JacobsZ19}, and to~\cite{Jacobs18c,Panangaden09} for more
general probabilistic semantics. We shall use the Bayesian network
from Section~\ref{sec:illustration} to illustrate the abstract
concepts that we introduce below.

\subsection{Distributions/states}\label{subsec:states}

In this context, we use the words `distribution' and `state'
interchangeably, for what is more precisely called a discrete
probability distribution, or also a monomial. A distribution on a set,
or sample space, $X$ is a formal convex combination of elements of
$X$, written as $r_{1}\ket{x_1} + \cdots + r_{n}\ket{x_n}$ with
$x_{i}\in X$ and $r_{i}\in [0,1]$ satisfying $\sum_{i} r_{i} = 1$. For
instance, the prior disease distribution in
Section~\ref{sec:illustration} can be written as $\frac{1}{100}\ket{d}
+ \frac{99}{100}\ket{\no{d}}$. This looks a bit heavy for a
distribution over a two-element set $\{d,\no{d}\}$, but this works
better for multiple elements, see
\textit{e.g.}~Example~\ref{ex:Halpern}. The ket notation $\ket{-}$ is
syntactic sugar that separates probabilities $r_i$ and elements $x_i$.

We shall write $\Dst(X)$ for the set of distributions (or states) on a
set $X$. We do not require that $X$ is finite itself, but $\Dst(X)$
contains only finite distributions. A distribution
$\sum_{i}r_{i}\ket{x_i} \in \Dst(X)$ may equivalently be described via
a probability mass function $\omega\colon X \rightarrow [0,1]$ with
finite support $\set{x}{\omega(x) \neq 0}$ and with $\sum_{x}\omega(x)
= 1$. We shall freely switch back-and-forth between formal convex sums
and probability mass functions.

\subsection{Channels}\label{subsec:channel}

A channel from a set $X$ to a set $Y$ is probabilistic computation
taking an element $x\in X$ as input and producing a distribution on
$Y$, indicating the probability of each output $y\in Y$. Thus, a
channel is a function $c\colon X \rightarrow \Dst(Y)$. We often write
it as $c\colon \kto{X}{Y}$, with a special arrow $\kto{}{}$. A channel
formalises a conditional probability $\Prob(y\mid x)$ as an actual
function $x \mapsto \Prob(y\mid x)$.  A channel thus captures an arrow
in a Bayesian network, namely as a stochastic matrix, or equivalently
as a conditional probability table.  For instance, the sensitivity
table in Figure~\ref{fig:diseaseBN} can be described as a channel:
\begin{equation}
\label{eqn:sensitivity}
\xymatrix{
\{d,\no{d}\}\ar[r]|-{\circ}^-{s} & \{t,\no{t}\}
\qquad\mbox{with}\qquad
{\left\{\begin{array}{rcl}
s(d) & = & \frac{9}{10}\ket{t} + \frac{1}{10}\ket{\no{t}}
\\
s(\no{d}) & = & \frac{1}{20}\ket{t} + \frac{19}{20}\ket{\no{t}}.
\end{array}\right.}
}
\end{equation}

\noindent This channel $s$ represents the arrow $\ovalbox{disease}
\rightarrow \ovalbox{test}$ in Figure~\ref{fig:diseaseBN} as a
probabilistic function $\kto{\{d,\no{d}\}}{\{t,\no{t}\}}$. Channels
provide a compositional semantics for Bayesian networks,
see~\cite{JacobsZ19} for more details.

Given a channel $c\colon \kto{X}{Y}$ one can transform a state
$\omega\in\Dst(X)$ on $X$ into a state $c \gg \omega \in\Dst(Y)$ on
$Y$. This corresponds to prediction. Concretely, we can
describe state transformation as below, first in mass function form,
and then as convex formal sum:
\[ \begin{array}{rclcrcl}
\big(c \gg \omega\big)(y)
& = &
\sum\limits_{x}\omega(x)\cdot c(x)(y)
& \;\mbox{ that is }\; &
c \gg \omega
& = &
\displaystyle\sum_{y} \textstyle\big(\sum\limits_{x}\omega(x)\cdot c(x)(y)\big)
   \bigket{y}.
\end{array} \]

\noindent For instance, the predicted test probability $\Prob(t) =
\frac{117}{2000}$ in Figure~\ref{fig:diseaseBN} can be obtained via
state transformation as:
\[ \begin{array}{rcl}
\lefteqn{\textstyle
   s \gg \Big(\frac{1}{100}\ket{d} + \frac{99}{100}\ket{\no{d}}\Big)}
\\
& = &
\big(\frac{1}{100}\cdot s(d)(t) + \frac{99}{100}\cdot s(\no{d})(t)\big)\ket{t}
+
\big(\frac{1}{100}\cdot s(d)(\no{t}) + 
   \frac{99}{100}\cdot s(\no{d})(\no{t})\big)\ket{\no{t}}
\\
& = &
\big(\frac{1}{100}\cdot \frac{9}{10} + 
   \frac{99}{100}\cdot \frac{1}{20}\big)\ket{t}
+
\big(\frac{1}{100}\cdot \frac{99}{100} + 
   \frac{99}{100}\cdot \frac{19}{20}\big)\ket{\no{t}}
\\
& = &
\frac{117}{2000}\ket{t} + \frac{1883}{2000}\ket{\no{t}}.
\end{array} \]

Given two channels $c\colon\kto{X}{Y}$ and $d\colon \kto{Y}{Z}$ we can
define their composite $d\klafter c \colon \kto{X}{Z}$ as:
\[ \begin{array}{rcccl}
\big(d \klafter c\big)(x)
& = &
d \gg c(x)
& = &
\displaystyle\sum_{z}\textstyle\big(\sum\limits_{y} c(x)(y)\cdot d(y)(z)\big)
   \bigket{z}.
\end{array} \]

\noindent There is a `Dirac' identity channel $\idmap \colon
\kto{X}{X}$ for this composition $\klafter$, with $\idmap(x) =
1\ket{x}$.  Moreover, $\klafter$ is associative and behaves well
wrt.\ state transformation: $(d \klafter c) \gg \omega = d \gg (c \gg
\omega)$. This gives an algebraic, compositional way for computing
probabilities --- especially in Bayesian networks.

Later on we shall use that each \emph{function} $f\colon X \rightarrow
Y$ can be turned into a `deterministic' channel $\hat{f} \colon X
\chanto Y$ via $\hat{f}(x) = 1\ket{f(x)}$. Then it is easy to see that
$\hat{g} \klafter \hat{f} = \widehat{g \after f}$.

For instance, the Bayesian network on the right in
Figure~\ref{fig:diseaseBN} involves an additional channel $e$ that
captures the certainty of the test evidence (for $r = \frac{8}{10}$):
\begin{equation}
\label{eqn:evidence}
\xymatrix{
\{t,\no{t}\}\ar[r]|-{\circ}^-{e} & \{c,\no{c}\}
\qquad\mbox{with}\qquad
{\left\{\begin{array}{rcl}
e(t) & = & \frac{8}{10}\ket{c} + \frac{2}{10}\ket{\no{c}}
\\
e(\no{t}) & = & \frac{2}{10}\ket{c} + \frac{8}{10}\ket{\no{c}}.
\end{array}\right.}
}
\end{equation}

\noindent We can now compute the predicted certainty $\Prob(c) =
\frac{4702}{20000}$, either via multiple state transformations, or via
a single state transformation of the composed channel $e \klafter s
\colon \kto{\{d,\no{d}\}}{\{c,\no{c}\}}$, in:
\[ \begin{array}{rcl}
\lefteqn{\textstyle(e \klafter s) \gg 
   \Big(\frac{1}{100}\ket{d} + \frac{99}{100}\ket{\no{d}}\Big)
\hspace*{\arraycolsep}=\hspace*{\arraycolsep}
e \gg \Big(s \gg 
   \Big(\frac{1}{100}\ket{d} + \frac{99}{100}\ket{\no{d}}\Big)\Big)}
\\
& = &
e \gg \Big(\frac{117}{2000}\ket{t} + \frac{1883}{2000}\ket{\no{t}}\Big)
\\
& = &
\big(\frac{117}{2000}\cdot \frac{8}{10} + 
   \frac{1883}{2000}\cdot \frac{2}{10}\big)\ket{c}
+
\big(\frac{117}{2000}\cdot \frac{2}{10} + 
   \frac{1883}{2000}\cdot \frac{8}{10}\big)\ket{\no{c}} \hspace*{5em}
\\
& = &
\frac{4702}{20000}\ket{c} + \frac{15298}{20000}\ket{\no{c}}.
\end{array} \]

\subsection{Predicates, validity and updating}\label{subsec:predicates}

For a distribution $\sigma\in\Dst(X)$ on $X$ and a predicate $p\colon
X \rightarrow [0,1]$ on $X$ we write $\sigma\models p$ for the
\emph{validity} of $p$ in $\sigma$. It can also be called the expected
value, since the definition is:
\begin{equation}
\label{eqn:validity}
\begin{array}{rcl}
\sigma\models p
& \;=\; &
{\displaystyle\sum}_{x}\, \sigma(x)\cdot p(x).
\end{array}
\end{equation}

\noindent For an event $E\subseteq X$ the validity
$\sigma\models\indic{E} \,= \sum_{x\in E}\omega(x)$ is usually written
as $\Prob(E)$, with the state $\sigma$ left implicit. We need a new
notation with the state $\sigma$ explicit, since the state is not
fixed: it changes through state transformation $\gg$.

If this validity $\sigma\models p$ is non-zero, we can define the
updated, conditioned distribution $\sigma|_{p}$ on $X$ as:
\begin{equation}
\label{eqn:conditioning}
\begin{array}{rclcrcl}
\sigma|_{p}(x)
& = &
\displaystyle\frac{\sigma(x)\cdot p(x)}{\sigma\models p}
& \qquad\mbox{that is}\qquad &
\sigma|_{p}
& = &
\displaystyle\sum_{x}\,\frac{\sigma(x)\cdot p(x)}{\sigma\models p}\bigket{x}.
\end{array}
\end{equation}

\noindent This updated/revised distribution $\sigma|_{p}$ is defined
quite generally, for fuzzy predicates $p$. It allows us to express the
usual form of conditional $\Prob(E\mid D)$ for events $E,D\subseteq X$
as $\sigma|_{\indic{D}} \models \indic{E}$.

The result below summarises some basic properties of updating with
fuzzy predicates, including Bayes' rule (in fuzzy form),
see~\cite{Jacobs18c,JacobsZ19}. It uses conjunction $p\andthen q$ of
two fuzzy predicates, defined as pointwise multiplication: $(p\andthen
q)(x) = p(x) \cdot q(x)$. There is an associated truth predicate $\one
\colon X \rightarrow [0,1]$ sending each element to $1$, that is,
$\one(x) = 1$. Then $p \andthen \one = p = \one \andthen p$. Moreover,
it uses that a fuzzy predicate $p$ can be multiplied with a scalar
$s\in [0,1]$ to $s\cdot p \colon X \rightarrow [0,1]$, namely via
$(s\cdot p)(x) = s\cdot p(x)$.

\begin{lemma}
\label{lem:conditioning}
Let $\sigma$ be distribution on a set $X$, and let $p,q$ be predicates
on $X$.
\begin{enumerate}
\item \label{lem:conditioning:bayes} Bayes' rule holds for (fuzzy) predicates:
\[ \begin{array}{rcccl}
\sigma|_{p}\models q
& \;=\; &
\displaystyle\frac{\sigma\models p \andthen q}{\sigma \models p}
& \;=\; &
\displaystyle\frac{(\sigma|_{q}\models p) \cdot (\sigma\models q)}
   {\sigma \models p}.
\end{array} \]

\item \label{lem:conditioning:and} Iterated conditionings commute:
\[ \begin{array}{rcccl}
\big(\sigma|_{p}\big)|_{q}
& = &
\sigma|_{p\andthen q}
& = &
\big(\sigma|_{q}\big)|_{p}.
\end{array} \]

\noindent Moreover, conditioning with truth has no effect:
$\sigma|_{\one} = \sigma$. 

\item \label{lem:conditioning:scalar} Conditioning is does not change
  when the predicate involved is multiplied with a non-zero scalar:
  $\omega|_{p} = \omega|_{s\cdot p}$.  \QED
\end{enumerate}
\end{lemma}

A basic property of updating $\omega|_{p}$ is that the validity
$\omega|_{p} \models p$ is greater than $\omega \models p$. Thus by
changing $\omega$ into $\omega|_{p}$ the predicate $p$ becomes `more
true' (see~\cite{Jacobs19b} for a proof and more details). That's why
we associate the phrase `improvement' with this form of updating
$\omega|_{p}$, which will be used below for Pearl's rule.

\subsection{Predicate transformation}\label{subsec:predtrans}

We have seen how a state $\omega\in\Dst(X)$ can be transformed in a
forward manner along a channel $c\colon \kto{X}{Y}$, to a state $c \gg
\omega$ on the codomain $Y$ of the channel. One can also transform
predicates along a channel, but in opposite direction: given a
predicate $q\colon Y \rightarrow [0,1]$, one obtains a predicate $c
\ll q \colon X \rightarrow [0,1]$ on the domain $X$ of the channel
via:
\[ \begin{array}{rclcrcl}
\big(c \ll q\big)(x)
& = &
\sum\limits_{y} c(x)(y)\cdot q(y).
\end{array} \]

\noindent One then easily checks that the validities $c \gg \omega
\models q$ and $\omega \models c \ll q$ are the same. Further there is
a compositionality result $(d \klafter c) \ll q = c \ll (d \ll q)$ so
that predicate transformation can be done by following the arrow /
channel structure of a Bayesian network in a step-by-step manner.

\subsection{Forward and backward inference}\label{subsec:inference}

We are now combining state transformation, predicate transformation,
and conditioning in order to identify two basic inference patterns,
namely forward inference and backward inference, see~\cite{JacobsZ16}
and~\cite{JacobsZ19}. We start from:
\[ \xymatrix{
\mbox{a state $\omega\in\Dst(X)$ on $X$, and a channel }\; 
   X\ar[r]|-{\circ}^-{c} & Y.
} \]

\begin{enumerate}
\item \textbf{Forward inference} with a predicate $p\colon X
  \rightarrow [0,1]$ is done by updating-and-state-transformation:
\[ c \gg \big(\omega|_{p}\big). \]

\noindent This yields a new distribution on $Y$.

\item \textbf{Backward inference} with a predicate $q\colon Y 
\rightarrow [0,1]$ is done by predicate-transformation-and-updating:
\[ \omega|_{c \ll q}. \]

\noindent This gives a new distribution on $X$.
\end{enumerate}

\noindent In the literature, see \textit{e.g.}~\cite{KollerF09},
forward inference is also called \emph{prediction} or \emph{causal
  reasoning}, and backward inference is called \emph{evidential
  reasoning} or \emph{explanation}.

In this context backward inference plays the more important role.  We
illustrate it for the disease-test example from
Section~\ref{sec:illustration}. Recall the characteristic function
$\indic{E} \colon X \rightarrow [0,1]$ associated with an event/subset
$E\subseteq X$. For an element $x\in X$ we simply write
$\indic{x}\colon X \rightarrow [0,1]$ instead of $\indic{\{x\}}$.

Let's write $\omega = \frac{1}{100}\ket{d} +
\frac{99}{100}\ket{\no{d}}$ for the prior disease probability from
Section~\ref{sec:illustration}. Updating it with positive test
evidence $\indic{t}$ happens via backward inference as $\omega|_{s \ll
  \indic{t}}$, using the sensitivity channel $s$
from~\eqref{eqn:sensitivity}. As illustration, we compute it
explicitly in several steps:
\[ \begin{array}{rcl}
(s \ll \indic{t})(d)
& = &
\sum_{x\in\{t,\no{t}\}} s(d)(x)\cdot \indic{t}(x)
\hspace*{\arraycolsep}=\hspace*{\arraycolsep}
s(d)(t)
\hspace*{\arraycolsep}=\hspace*{\arraycolsep}
\frac{9}{10}
\\
(s \ll \indic{t})(\no{d})
& = &
s(\no{d})(t)
\hspace*{\arraycolsep}=\hspace*{\arraycolsep}
\frac{1}{20}
\\
\omega \models s \ll \indic{t}
& = &
\sum_{x\in\{d,\no{d}\}} \omega(x) \cdot (s \ll \indic{t})(x)
\hspace*{\arraycolsep}=\hspace*{\arraycolsep}
\frac{1}{100}\cdot \frac{9}{10} + \frac{99}{100}\cdot \frac{1}{20}
\hspace*{\arraycolsep}=\hspace*{\arraycolsep}
\frac{117}{2000}
\\
\omega|_{s \ll \indic{t}}
& = &
\sum_{x\in\{d,\no{d}\}} \displaystyle\frac{\omega(x)\cdot (s \ll \indic{t})(x)}
   {\omega \models s \ll \indic{t}}\bigket{x}
\\[+0.8em]
& = &
\displaystyle
\frac{\nicefrac{1}{100}\cdot\nicefrac{9}{10}}{\nicefrac{117}{2000}}\ket{d} +
\frac{\nicefrac{99}{100}\cdot\nicefrac{1}{20}}{\nicefrac{117}{2000}}\ket{\no{d}}
\hspace*{\arraycolsep}=\hspace*{\arraycolsep}
\textstyle\frac{117}{2000}\ket{d} + \frac{1883}{2000}\ket{\no{d}}.
\end{array} \]

\noindent We see how the probability $\Prob(d\mid t) =
\frac{117}{2000}$ from Section~\ref{sec:illustration} re-emerges, via
a channel-based computation. In a similar way one can compute
$\Prob(d\mid c) = \frac{148}{4702}$ via backward inference as:
\[ \begin{array}{rcccl}
\omega|_{(e \klafter s) \ll \indic{c}}
& = &
\omega|_{s \ll (e \ll \indic{c})}
& = &
\frac{148}{4702}\ket{d} + \frac{4554}{4702}\ket{\no{d}}.
\end{array} \]

\noindent We see that backward inference can be done in a
compositional manner, following the graph structure $\ovalbox{disease}
\rightarrow \ovalbox{test} \rightarrow \ovalbox{certainty}$ on the
right in Figure~\ref{fig:diseaseBN}.

\subsection{Pearl's update rule}\label{subsec:constructive}

We are now finally in a position to describe Pearl's rule of virtual
evidence, that is. Let's write $2 = \{0,1\}$ for a generic two-element
set. The crucial observation is that extending a Bayesian network at
node $X$ with virtual evidence of the form $X \chanto 2$ corresponds
to introducing a fuzzy predicate for updating. This works since
$\Dst(2) \cong [0,1]$, so that a table/channel $X \chanto 2$ to a
binary node $2$ corresponds to a fuzzy predicate $X \rightarrow
[0,1]$. Indeed, the soft evidence described in
Section~\ref{sec:illustration} can be captured by a fuzzy predicate $p
\colon \{t, \no{t}\} \rightarrow [0,1]$ with $p(t) = \frac{8}{10}$ and
$p(\no{t}) = \frac{2}{10}$. Pearl's rule then amounts to backward
reasoning of the form $\omega|_{s \ll p} = \frac{148}{4702}\ket{d} +
\frac{4554}{4702}\ket{\no{d}}$, as computed above. This works since $p
= e \ll \indic{c}$.

We now formalise Pearl's rule in a channel-based setting.

\begin{definition}
\label{def:constructive}
Let $c\colon X \chanto Y$ be a channel with prior $\sigma\in\Dst(X)$.
Given a predicate $q\colon Y \rightarrow [0,1]$ on the channel's
codomain $Y$, Pearl's rule uses backward inference to update the prior
$\sigma$ to the posterior:
\[ \sigma|_{c \ll q} \,\in\, \Dst(X). \]
\end{definition}

This formulation of Pearl's rule does not refer to any extension of a
Bayesian network with a binary node. Still, one may consider the
channel $c\colon X \chanto Y$ as a mini-network that is extended with
predicate $q$, as in: $\smash{\ovalbox{$X$}
  \stackrel{c}{\longrightarrow} \ovalbox{$Y$}
  \stackrel{q}{\longrightarrow} \ovalbox{2}}$.

We finish this section with some basic properties. They follow easily
from Lemma~\ref{lem:conditioning}.

\begin{proposition}
\label{prop:constructive}
Let $\omega\in\Dst(X)$ be a distribution and $c\colon X \chanto Y$
be a channel.
\begin{enumerate}
\item \label{prop:constructive:scalar} Backward inference is
  invariant under pointwise/scalar multiplication of the evidence
  predicate with a non-zero probability $s\in(0,1]$,
\[ \begin{array}{rcl}
\omega|_{c \ll (s\cdot p)}
& = &
\omega|_{c \ll p}.
\end{array} \]

\item \label{prop:constructive:effect} Backward inference with a
  non-zero constant (uniform) predicate $s\cdot \one$ as evidence has
  no effect:
\[ \begin{array}{rcl}
\omega|_{c \ll (s\cdot \one)}
& = &
\omega.
\end{array} \]

\item \label{prop:constructive:and} Iterated applications of backward
  inference commute, and satisfy:
\[ \begin{array}{rcccccl}
\omega\big|_{c \ll p}\big|_{d \ll q}
& = &
\omega\big|_{(c \ll p) \andthen (d \ll q)}
& = &
\omega\big|_{d \ll q}\big|_{c \ll p}.
\end{array} \eqno{\QEDbox} \]
\end{enumerate}
\end{proposition}

\section{Bayesian inversion and Jeffrey's update rule}\label{sec:inversion}

One way to read Bayes' rule is as an `inversion' property, turning a
conditional probability $\Prob(y\mid x)$ into $\Prob(x\mid y)$. Since
channels correspond to conditional probabilities, such inversion can
be formulated for channels as well, see~\cite{ClercDDG17} and
also~\cite{ChoJ19}. This inversion is relevant because it allows us to
give a precise description of Jeffrey's rule.

\subsection{Bayesian inversion via updating}\label{subsec:inversion}

Let $c\colon X \chanto Y$ be a channel, with a prior
distribution/state $\sigma\in\Dst(X)$ on its domain. In this
situation, with a certain side-condition fulfilled, we can define an
inverted channel $c^{\dag}_{\sigma} \colon Y \chanto X$ in the
opposite direction. This function $c^{\dag}_{\sigma} \colon Y
\rightarrow \Dst(X)$ is defined via backward inference with point
predicates $\indic{y} \colon Y \rightarrow [0,1]$, for $y\in Y$.
\begin{equation}
\label{eqn:inversion}
\begin{array}{rcccl}
c^{\dag}_{\sigma}(y)
& = &
\sigma|_{c \ll \indic{y}}
& = &
\displaystyle\sum_{x}\,
   \frac{\sigma(x)\cdot c(x)(y)}{(c \gg \sigma)(y)}\bigket{x}.
\end{array}
\end{equation}

\noindent The distribution $c^{\dag}_{\sigma}(y) \in \Dst(X)$ is the
posterior, obtained after observing $y\in Y$, that is, after updating
with point evidence $\indic{y}$. This definition only makes sense if
the transformed state $c \gg \sigma$ has full support, that is, if $(c
\gg \sigma)(y) \neq 0$ for each $y\in Y$.

The dagger notation $c^{\dag}_{\sigma}$ for probabilistic computations
comes from~\cite{ClercDDG17}; the subscript $\sigma$ may be omitted if
it is clear from the context. This dagger satisfies some basic
algebraic properties: inverting twice yields the original channel:
$(c^{\dag})^{\dag} = c$. Moreover, inversion interacts appropriately
with channel composition: $(d \klafter c)^{\dag} = c^{\dag} \klafter
d^{\dag}$.  The dagger notation is more common in quantum theory,
where unitary computations are reversible, and has been formalised in
terms of dagger categories, see \textit{e.g.}~\cite{CoeckeK16}.


Given the above definition~\eqref{eqn:inversion} we see that we have
implicitly already computed the Bayesian inversion $s^{\dag}$ of the
sensitivity channel $s \colon \{d,\no{d}\} \chanto \{t,\no{t}\}$ from
Section~\ref{sec:illustration}, namely via the conditional
probabilities $\Prob(d\mid t) = \frac{18}{117}$ and $\Prob(d\mid \no{t}) =
\frac{2}{1883}$. Thus we have:
\begin{equation}
\label{eqn:sensitivitydagger}
\xymatrix{
\{t,\no{t}\}\ar[r]|-{\circ}^-{s^{\dag}} & \{d,\no{d}\}
\qquad\mbox{with}\qquad
{\left\{\begin{array}{rcl}
s^{\dag}(t) & = & \frac{18}{117}\ket{d} + \frac{99}{117}\ket{\no{d}}
\\
s^{\dag}(\no{t}) & = & \frac{2}{1883}\ket{d} + \frac{1881}{1883}\ket{\no{d}}.
\end{array}\right.}
}
\end{equation}

\subsection{Bayesian inversion and inference}\label{subsec:inversioninference}

In Subsection~\ref{subsec:inference} we have described forward and
backward inference along a channel. It turns out that forward becomes
backward --- and vice-versa --- when we use an inverted channel.  This
illustrates that the basic notions of inference, transformation and
inversion are mathematically closely related. The proof is obtained by
unwrapping the relevant definitions and is left to the interested
reader.

\begin{theorem}
\label{thm:daggerconditioning}
Let $c\colon \kto{X}{Y}$ be a channel with a state $\sigma\in\Dst(X)$ on
its domain, such that $\tau = c \gg \sigma$ has full support.
\begin{enumerate}
\item \label{thm:daggerconditioning:back} Given a predicate $q$ on
  $Y$, we can express backward inference along $c$ as forward
  inference along $c^{\dag}$ via:
\[ \begin{array}{rcl}
\sigma|_{c \ll q}
& = &
c^{\dag} \gg \big(\tau|_{q}\big).
\end{array} \]


\item \label{thm:daggerconditioning:forward} Given a predicate $p$ on
  $X$, we can express forward inference along $c$ as backward
  inference along $c^{\dag}$:
\[ \begin{array}{rcl}
c \gg \big(\sigma|_{p}\big)
& = &
\tau|_{c^{\dag} \ll p}.
\end{array} \eqno{\QEDbox} \]
\end{enumerate}
\end{theorem}

\auxproof{
\begin{myproof}
\begin{enumerate}
\item $\begin{array}[t]{rcl}
\Big(c^{\dag}_{\sigma} \gg \big((c \gg \sigma)|_{q}\big)\Big)(x)
& = &
\sum_{y} c^{\dag}_{\sigma}(y)(x) \cdot (c \gg \sigma)|_{q}(y)
\\[+0.3em]
& = &
\sum_{y} \displaystyle\frac{c(x)(y)\cdot\sigma(x)}{(c \gg \sigma)(y)} \cdot 
    \frac{(c \gg \sigma)(y) \cdot q(y)}{c \gg \sigma \models q}
\\[+0.8em]
& = &
\displaystyle\frac{\sigma(x)\cdot \big(\sum_{y} c(x)(y)\cdot q(y)\big)}
   {\sigma \models c \ll q}
\\[+0.8em]
& = &
\displaystyle\frac{\sigma(x)\cdot (c \ll q)(x)}
   {\sigma \models c \ll q}
\\
& = &
\sigma|_{c \ll q}(x).
\end{array}$

\item Similarly,
\[ \begin{array}[b]{rcl}
\Big(\big(c \gg \sigma\big)\big|_{c_{\sigma}^{\dag} \ll p}\Big)(y)
& = &
\displaystyle\frac{(c \gg \sigma)(y) \cdot (c_{\sigma}^{\dag} \ll p)(y)}
   {c \gg \sigma \models c_{\sigma}^{\dag} \ll p}
\\[+0.8em]
& = &
\sum_{x}\displaystyle\frac{(c \gg \sigma)(y) 
   \cdot c_{\sigma}^{\dag}(y)(x) \cdot p(x)}
   {c_{\sigma}^{\dag} \gg (c \gg \sigma) \models p}
\\[+1em]
& = &
\sum_{x}\displaystyle\frac{(c \gg \sigma)(y) 
   \cdot \frac{\sigma(x) \cdot c(x)(y)}{(c \gg \sigma)(y)} \cdot p(x)}
   {\sigma \models p} \quad \mbox{by point~\eqref{thm:daggerconditioning:back}}
\\[+0.8em]
& = &
\sum_{x} c(x)(y)\cdot \displaystyle\frac{\sigma(x)\cdot p(x)}{\sigma\models p}
\\
& = &
\Big(c \gg (\sigma|_{p})\Big)(y).
\end{array} \eqno{\QEDbox} \]
\end{enumerate}
\end{myproof}
}

\subsection{Jeffrey's update rule}\label{subsec:destructive}

At this stage we have prepared the grounds to give a channel-based
formulation of Jeffrey's rule. It uses the inversion of a channel for
backtracking.

\begin{definition}
\label{def:destructive}
Let $c\colon X \chanto Y$ be a channel with prior $\sigma\in\Dst(X)$.
Given a state $\rho\in\Dst(Y)$ on the channel's codomain $Y$,
Jeffrey's rule involves using state transformation along the inverted
channel $c^{\dag}_{\sigma}$ to update the prior $\sigma$ to the
posterior:
\[ c^{\dag}_{\sigma} \gg \rho \,\in\, \Dst(X). \]
\end{definition}

Indeed, this state transformation is what we have used to compute
Jeffrey's update in Section~\ref{sec:illustration} as convex
combination~\eqref{eqn:disease:DU}. More explicitly, translating
$80\%$ certainty into a state, and using $s^{\dag}$
from~\eqref{eqn:sensitivitydagger} we get approximately $12\%$ disease
likelihood via:
\[ \begin{array}{rcl}
s^{\dag} \gg \big(\frac{8}{10}\ket{t} + \frac{2}{10}\ket{\no{t}}\big)
& = &
\big(\frac{8}{10}\cdot\frac{18}{117} + 
   \frac{2}{10}\cdot\frac{2}{1883}\big)\ket{d} +
\big(\frac{8}{10}\cdot\frac{99}{117} + 
   \frac{2}{10}\cdot\frac{1881}{1883}\big)\ket{\no{d}}
\\
& = &
\frac{27162}{220311}\ket{d} + \frac{193149}{220311}\ket{\no{d}}.
\end{array} \]

We continue with some properties of Jeffrey's updating. The
translations back-and-forth between the Pearl's and Jeffrey's rules
are due to~\cite{ChanD05}; they are translated here to the current
setting

\begin{proposition}
\label{prop:destructive}
Let $c\colon \kto{X}{Y}$ be a channel with a state $\sigma\in\Dst(X)$ on
its domain, such that $\tau = c \gg \sigma$ has full support.
\begin{enumerate}
\item \label{prop:destructive:effect} Jeffrey's updating with the
  predicted state $\tau = c \gg \sigma$ does not have any effect:
\[ \begin{array}{rcl}
c^{\dag} \gg \tau
& = &
\sigma.
\end{array} \]

\item \label{prop:destructive:and} Successive Jeffrey updates do not
  commute: given evidence $\rho_{1}, \rho_{2} \in \Dst(Y)$, giving
  $\sigma_{i} = c^{\dag}_{\sigma} \gg \rho_{i} \in \Dst(X)$, then, in
  general,
\[ \begin{array}{rcl}
c^{\dag}_{\sigma_1} \gg \rho_{2}
& \neq &
c^{\dag}_{\sigma_2} \gg \rho_{1}.
\end{array} \]

\item \label{prop:destructive:point} Jeffrey's and Pearl's updating
  coincide on point evidence:
\[ \begin{array}{rcccl}
c^{\dag}_{\sigma} \gg 1\ket{y}
& = &
c^{\dag}_{\sigma}(y)
& = &
\sigma|_{c \ll \indic{y}}.
\end{array} \]

\item Pearl's updating can be expressed as Jeffrey's updating,
  by turning predicate evidence $q\colon Y \rightarrow [0,1]$ into
  state evidence $\tau|_{q}$, see again
  Theorem~\ref{thm:daggerconditioning}~\eqref{thm:daggerconditioning:back}.

\item Jeffrey's updating can also be expressed as Pearl's updating:
  for a state $\rho\in\Dst(Y)$ write $\nicefrac{\rho}{\tau}$ for the
  predicate $y \mapsto \frac{\rho(y)}{\tau(y)}$, suitably rescaled to
  $[0,1]$ if needed; then:
\[ \begin{array}{rcl}
c^{\dag}_{\sigma} \gg \rho
& = &
\sigma|_{c \ll \nicefrac{\rho}{\tau}}.
\end{array} \]
\end{enumerate}
\end{proposition}

\begin{myproof}
Only the last point is non-trivial. First we note that $\tau \models
\nicefrac{\rho}{\tau} = 1$, since $\rho$ is a state:
\[ \begin{array}{rcccccl}
\tau \models \nicefrac{\rho}{\tau}
& = &
\sum_{y} \tau(y) \cdot \frac{\rho(y)}{\tau(y)}
& = &
\sum_{y} \rho(y)
& = &
1.
\end{array} \]

\noindent But then, for $x\in X$,
\[ \begin{array}[b]{rcl}
\big(c^{\dag}_{\sigma} \gg \rho\big)(x)
\hspace*{\arraycolsep}=\hspace*{\arraycolsep}
\sum_{y} \rho(y) \cdot c^{\dag}_{\sigma}(y)(x)
& \smash{\stackrel{\eqref{eqn:inversion}}{=}} &
\sum_{y} \rho(y) \cdot \displaystyle\frac{\sigma(x)\cdot c(x)(y)}{\tau(y)}
\\
& = &
\sigma(x) \cdot \sum_{y} c(x)(y) \cdot \nicefrac{\rho}{\tau}(y)
\\[+0.5em]
& = &
\displaystyle\frac{\sigma(x) \cdot (c \ll \nicefrac{\rho}{\tau})(x)}
   {c \gg \sigma \models \nicefrac{\rho}{\tau}}
   \quad \mbox{ as just shown}
\\[+1em]
& = &
\displaystyle\frac{\sigma(x) \cdot (c \ll \nicefrac{\rho}{\tau})(x)}
   {\sigma \models c \ll \nicefrac{\rho}{\tau}}
\\
& \smash{\stackrel{\eqref{eqn:conditioning}}{=}}\hspace*{\arraycolsep} &
\big(\sigma|_{c \ll \nicefrac{\rho}{\tau}}\big)(x).
\end{array} \eqno{\QEDbox} \]
\end{myproof}

We conclude this section with a couple of remarks.

\begin{remark}
\label{rem:destructive}
\begin{enumerate}
\item \label{rem:destructive:terminology}
  Proposition~\ref{prop:destructive}~\eqref{prop:destructive:effect}
  shows that $\sigma = c^{\dag}_{\sigma} \gg (c \gg \sigma)$. This
  means that a state $\sigma$ can be reconstructed, via Jeffrey's
  updating, from what we can predict, namely from $c \gg \sigma$. At
  this same time it shows that in Jeffrey's updating
  $c^{\dag}_{\sigma} \gg \rho$ the `state of affairs' $\rho$ that we
  encounter as evidence replaces the prediction $c\gg\sigma$, where
  the inversion $c^{\dag}_{\sigma}$ is used for back-tracking. This
  replacement, of $c \gg \sigma$ by $\rho$, is where the `shock' or
  `surpise' of Jeffrey's rule can be located. We also use the terms
  `correction' and `adjustment' for this process, see
  Table~\eqref{table:naming}.

\item \label{rem:destructive:statepred} We briefly come back to the
  issue whether softness/uncertainty should be represented as a state
  or as a predicate. Pragmatically, one can go either way, since each
  state is a predicate, and in the other direction a predicate (on a
  finite) set can be normalised to a state, and the scaling factor
  involved does not affect the outcome in conditioning, see
  Lemma~\ref{lem:conditioning}~\eqref{lem:conditioning:scalar}.

From a structural, algebraic perspective however, there are
significant differences between states and predicates. For one, they
form different mathematical structures: states are convex sets,
whereas predicates are effect modules with a monoid structure (for
conjunction), see \textit{e.g.}~\cite{Jacobs15d,Jacobs18c} for
details. This means that they come with different algebraic
operations. For instance, predicates are closed under scalar
multiplication, but states are not.  In addition, there are different
transformation operations: states can be transformed forwardly $\gg$
along a channel, and predicates backwardly $\ll$. These operations are
mathematically well-behaved: convex combinations of states are
preserved by state transformation, whereas the effect module structure
is preserved by predicate transformation. States and predicates are
dual to each other, see \textit{e.g.}~\cite{Jacobs17b,Jacobs18c} for a
wider perspective. These structural differences suffice to keep states
and predicates apart in a mathematically precise manner.

In addition, the mathematical distinction between states and
predicates fits the terminological distinction of
Table~\eqref{table:naming}: a state of affairs in Jeffrey's updating
corresponds to a state / probability distribution, whereas (soft)
evidence corresponds to a (fuzzy) predicate. This means that the
terminology has a mathematical basis.
\end{enumerate}
\end{remark}

\section{Literature review}\label{sec:literature}

This section compares the channel-based explanation of
Jeffrey's/Pearl's updating of the previous two sections, together with
its informal interpretation of Section~\ref{sec:observations}, to some
relevant material in the literature. It first reviews some examples
and then looks at earlier approaches to soft evidence that focus
mainly on how to formulate softness in the first place.

\subsection{Examples from the literature}\label{subsec:literatureexamples}

\begin{example}
\label{ex:Halpern}
First we consider the following question
from~\cite[Example~3.10.1]{Halpern03}.
\begin{quote}
Suppose that an object is either red ($r$), blue ($b$), green ($g$),
or yellow ($y$). An agent initially ascribes probability
$\nicefrac{1}{5}$ to each of red, blue, and green, and probability
$\nicefrac{2}{5}$ to yellow. Then the agent gets a quick glimpse of
the object in a dimly lit room. As a result of this glimpse, he
believes that the object is probably a darker color, although he is
not sure. He thus ascribes probability $.7$ to it being green or blue
and probability $.3$ to it being red or yellow. How should he update
his initial probability measure based on this observation?
\end{quote}

\noindent The prior probability distribution is in this case $\sigma =
\frac{1}{5}\ket{r} + \frac{1}{5}\ket{b} + \frac{1}{5}\ket{g} +
\frac{2}{5}\ket{y}$. We see that the colors in this example are
partitioned in two combinations, namely `green or blue' and `red or
yellow'. We capture this via a two-element set $\{gb, ry\}$. There is
then an obvious (deterministic) channel:
\[ \xymatrix{
\{r,b,g,y\}\ar[r]|-{\circ}^-{c} & \{gb,ry\}
\qquad\mbox{with}\qquad
{\left\{\begin{array}{rcl}
c(r) & = & 1\ket{ry}
\\
c(b) & = & 1\ket{gb}
\\
c(g) & = & 1\ket{gb}
\\
c(y) & = & 1\ket{ry}
\end{array}\right.}
} \]

\noindent The above quote does not suggest whether the new information
should be used for correction, or for improvement. In~\cite{Halpern03}
the first approach is chosen.  Here we elaborate both.

The posterior (updated) probability distribution, computed via
Jeffrey's rule, is obtained by doing state transformation with the
inverted channel and the `glimpse' as state of affairs:
\[ \begin{array}{rcl}
c^{\dag}_{\sigma} \gg \big(\frac{7}{10}\ket{gb} + \frac{3}{10}\ket{ry}\big)
& = &
\frac{1}{10}\ket{r} + \frac{7}{20}\ket{b} + 
    \frac{7}{20}\ket{g} + \frac{1}{5}\ket{y}.
\end{array} \]

\noindent However, one can also translate the `glimpse' into a fuzzy
predicate $p\colon \{gb, ry\} \rightarrow [0,1]$ with $p(gb) =
\frac{7}{10}$ and $p(ry) = \frac{3}{10}$. Pearl's update rule
then gives a different outcome:
\[ \begin{array}{rcl}
\sigma|_{c \ll p}
& = &
\frac{3}{23}\ket{r} + \frac{7}{23}\ket{b} + 
    \frac{7}{23}\ket{g} + \frac{6}{23}\ket{y}.
\end{array} \]
\end{example}

This example is an instance of a frequently occurring setting in which
Jeffrey's rule is formulated (notably in~\cite{Halpern03}, to which we
refer for details), namely when the channel involved is deterministic.
Consider a function $f\colon X \rightarrow I$, giving a partition of
the set $X$ via subsets $U_{i} = f^{-1}(i) = \setin{x}{X}{f(x) =
  i}$. This function can be turned into a `deterministic' channel
$\hat{f} \colon X \chanto I$, via $\hat{f}(x) = 1\ket{f(x)}$.

\begin{lemma}
\label{lem:deterministic}
Let $f\colon X \rightarrow I$ be a function/partition, to be used as
deterministic channel, as just described, together with a prior
$\omega\in\Dst(X)$. Applying Jeffrey's rule to a new state of affairs
$\rho\in\Dst(I)$ gives as posterior:
\begin{equation}
\label{eqn:deterministic}
\begin{array}{rclcrcl}
\hat{f}^{\dag}_{\omega} \gg \rho
& = &
\sum_{i}\, \rho(i)\cdot\omega|_{\indic{U_i}}
& \qquad\mbox{satisfying}\qquad &
\hat{f} \gg \big(\hat{f}^{\dag}_{\omega} \gg \rho\big)
& = &
\rho.
\end{array}
\end{equation}

\noindent Moreover, wrt.\ the total variation distance function $d$ one has:
\[ \begin{array}{rcl}
d\big(\hat{f}^{\dag}_{\omega} \gg \rho, \omega)
& = &
\displaystyle\bigwedge\set{d(\omega,\omega')}{\omega'\in\Dst(X)
   \mbox{ with }\hat{f} \gg \omega' = \rho}.
\end{array}\eqno{\QEDbox} \]
\end{lemma}

\auxproof{
\[ \begin{array}{rcl}
\big(\hat{f}^{\dag}_{\omega} \gg \rho\big)(x)
& = &
\sum_{i} \rho(i) \cdot \hat{f}^{\dag}_{\omega}(i)(x)
\\
& = &
\sum_{i} \rho(i) \cdot \displaystyle\frac{\omega(x) \cdot \hat{f}(x)(i)}
   {(\hat{f} \gg \omega)(i)}
\\
& = &
\sum_{i}\rho(i) \cdot \left\{\begin{array}{ll}
\displaystyle\frac{\omega(x)}{\sum_{z\in f^{-1}(i)}\omega(z)} \quad
   & \mbox{if }x\in U_{i}
\\
0 & \mbox{otherwise}
\end{array}\right.
\\
& = &
\sum_{i}\rho(i) \cdot \left\{\begin{array}{ll}
\omega|_{\indic{U_i}}(x) \quad & \mbox{if }x\in U_{i}
\\
0 & \mbox{otherwise}
\end{array}\right.
\\
& = &
\sum_{i} \rho(i)\cdot\omega|_{\indic{U_i}}(x)
\\
\big(\hat{f} \gg \big(\hat{f}^{\dag}_{\omega} \gg \rho\big)\big)(i)
& = &
\sum_{x\in f^{-1}(i)} \big(\hat{f}^{\dag}_{\omega} \gg \rho\big)(x)
\\
& = &
\sum_{x\in U_{i}} \rho(i) \cdot \omega|_{\indic{U_i}}(x)
\\[+0.5em]
& = &
\rho(i) \cdot \displaystyle
   \frac{\sum_{x\in U_{i}} \omega(x) \cdot \indic{U_i}(x)}
   {\omega\models\indic{U_i}}
\\[+1em]
& = &
\rho(i) \cdot \displaystyle
   \frac{\omega\models\indic{U_i}}{\omega\models\indic{U_i}}
\\
& = &
\rho(i).
\end{array} \]

An associated property that is often mentioned (\textit{e.g.}
in~\cite{Halpern03}) in relation to this deterministic formulation of
Jeffrey's rule concerns the (total variation) distance:
\[ \begin{array}{rcl}
d\big(\hat{f}^{\dag}_{\omega} \gg \rho, \omega)
& = &
\displaystyle\bigwedge\set{d(\omega',\omega)}{\omega'\in\Dst(X)
   \mbox{ with }\hat{f} \gg \omega' = \rho}
\end{array} \]

\noindent Since $\hat{f} \gg \big(\hat{f}^{\dag}_{\omega} \gg
\rho\big) = \rho$, as we have just seen, $(\geq)$ holds. For $(\leq)$,
let $\omega'\in\Dst(X)$ satisfy $\hat{f} \gg \omega' = \rho$. Then:
\[ \begin{array}{rcl}
d\big(\hat{f}^{\dag}_{\omega} \gg \rho, \omega)
& = &
\sum_{x} \big|\big(\hat{f}^{\dag}_{\omega} \gg \rho\big)(x) - \omega(x)\big|
\\
& = &
\sum_{i} \sum_{x\in U_{i}} \big|\rho(i) \cdot \omega|_{\indic{U_i}}(x) - \omega(x)\big|
\\
& = &
\sum_{i} \sum_{x\in U_{i}} \big|
   \frac{\rho(i)\cdot\omega(x)}{\omega \models \indic{U_i}} - \omega(x)\big|
\\
& = &
\sum_{i} \sum_{x\in U_{i}} \omega(x) \cdot \big|
   \frac{\rho(i)}{\omega \models \indic{U_i}} - 1\big|
\\
& = &
\sum_{i} (\omega \models \indic{U_i}) \cdot \big|
   \frac{\rho(i)}{\omega \models \indic{U_i}} - 1\big|
\\
& = &
\sum_{i} \big|\rho(i) - (\omega \models \indic{U_i})\big|
\\
& = &
\sum_{i} \big|(\hat{f} \gg \omega')(i) - (\omega \models \indic{U_i})\big|
\\
& = &
\sum_{i} \big|(\sum_{x\in U_i}\omega'(x)) - (\omega \models \indic{U_i})\big|
\\
& = &
\sum_{i} \big|(\omega' \models \indic{U_i}) - (\omega \models \indic{U_i})\big|
\\
& \leq &
\sum_{U\subseteq X} \big|(\omega' \models \indic{U}) - 
   (\omega \models \indic{U})\big|
\\
& = &
d(\omega', \omega).
\end{array} \]
}

The equation on the left in~\eqref{eqn:deterministic} describes
Jeffrey's update as a convex combination of updated states
$\omega|_{\indic{U_i}}$, conditioned to the partitions $U_i$, with
probabilities $\rho(i)$. The equation on the right
in~\eqref{eqn:deterministic} illustrates the `destructive' character
of Jeffrey's rule: the prediction after the update is equal to new
situation: the original prediction $\hat{f} \gg \omega$ is simply
overridden by $\rho$. The equations in this lemma hold because
$\hat{f}$ is a deterministic channel and do not hold for arbitrary
channels. Since many early examples of Jeffrey's rule involve such
partitions via deterministic channels, where the effect / predicted
state $\hat{f} \gg \omega'$ of the updated state $\omega' =
\hat{f}_{\omega}^{\,\dag} \gg \rho$ is equal to the uncertain evidence
$\rho$, \textit{i.e.}~$\hat{f} \gg \omega' = \rho$, the idea emerged
that for Jeffrey's rule the specification of the evidence $\rho$ must
happen in terms of the effect $\hat{f} \gg \omega'$. But, as said,
this only works for deterministic channels, not in general. We return
to this point below, in point~1 in Subsection~\ref{subsec:specmethod}.

For a general, not-deterministic channel $c\colon X \chanto Y$ with
prior state $\omega\in\Dst(X)$ and evidence state $\rho\in\Dst(Y)$ one
can prove:
\[ \begin{array}{rcl}
d\big(c^{\dag}_{\omega} \gg \rho, \omega)
& \leq &
\displaystyle\bigwedge_{\omega'\in\Dst(X)} 
   d(\omega,\omega') + d(c \gg \omega', \rho).
\end{array} \]

\auxproof{
We first recall that state transformation is non-expansive:
\[ \begin{array}{rcl}
d(c \gg \sigma, c \gg \sigma')
& = &
\sum_{y} \big|\, (c\gg\sigma)(y) - (c\gg\sigma')(y)\,\big|
\\
& = &
\sum_{y} \big|\, \big(\sum_{x} \sigma(x)\cdot c(x)(y)\big) - 
   \big(\sum_{x} \sigma(x)\cdot c(x)(y)\big) \,\big|
\\
& = &
\sum_{y} \big|\, \sum_{x} c(x)(y) \cdot \big(\sigma(x) - \sigma'(x)\big) \,\big|
\\
& \leq &
\sum_{x,y} c(x)(y) \cdot \big|\, \sigma(x) - \sigma'(x) \,\big|
\\
& = &
\sum_{x} \big(\sum_{y} c(x)(y)\big) \cdot \big|\, \sigma(x) - \sigma'(x) \,\big|
\\
& = &
\sum_{x} \big|\, \sigma(x) - \sigma'(x) \,\big|
\\
& = &
d(\sigma, \sigma').
\end{array} \]

\noindent For an arbitrary $\omega'\in\Dst(X)$ we have:
\[ \begin{array}{rcl}
d(\omega,\omega') + d(c \gg \omega', \rho)
& \geq &
d(c \gg \omega, c\gg \omega') + d(c \gg \omega', \rho)
\\
& \geq &
d(c \gg \omega, \rho)
\\
& = &
d(c^{\dag}_{\omega} \gg (c \gg \omega), c^{\dag}_{\omega} \gg \rho)
\\
& = &
d(\omega, c^{\dag}_{\omega} \gg \rho).
\end{array} \]
}

\begin{example}
\label{ex:Barber}
We turn to the following Bayesian network.
\[ \xymatrix@C-1pc@R-1pc{
{\setlength\tabcolsep{0.2em}
   \renewcommand{\arraystretch}{1}
\begin{tabular}{|c|}
\hline
$\Prob$(burglar) \\
\hline\hline
$0.01$ \\
\hline
\end{tabular}}
& \ovalbox{\strut burglar}\ar[ddr] & & 
{\hspace*{-1em}\ovalbox{\strut earthquake}\hspace*{-1em}}\ar[ddl]
   \rlap{\qquad\smash{\setlength\tabcolsep{0.2em}\renewcommand{\arraystretch}{1}
\begin{tabular}{|c|}
\hline
$\Prob$(earthquake) \\
\hline\hline
$0.000001$ \\
\hline
\end{tabular}}} 
\\
\\
& & \ovalbox{\strut alarm}
  \rlap{\hspace*{2em}\smash{\setlength\tabcolsep{0.2em}
\renewcommand{\arraystretch}{1}
\begin{tabular}{|c|c|c|}
\hline
burglar & earthquake & $\Prob$(alarm) \\
\hline\hline
$b$ & $e$ & $0.9999$ \\
\hline
$b$ & $\no{e}$ & $0.99$ \\
\hline
$\no{b}$ & $e$ & $0.99$ \\
\hline
$\no{b}$ & $\no{e}$ & $0.0001$ \\
\hline
\end{tabular}}}
& &
}\hspace*{10em} \]

\vspace*{2em}

\noindent The a prori probabilities of a burglary and an earthquake,
given in the upper tables, can be written as probability distributions:
\[ \begin{array}{rclcrcl}
\omega
& = &
0.01\ket{b} + 0.99\ket{\no{b}}
& \qquad\mbox{and}\qquad &
\sigma
& = &
0.000001\ket{e} + 0.999999\ket{\no{e}}.
\end{array} \]

\noindent These two states can be combined to a `joint' product state
on the product space $\{b,\no{b}\} \times \{e,\no{e}\}$, written as:
\[ \begin{array}{rcl}
\sigma \otimes \omega
& = &
0.00000001\ket{b,e} + 0.00999999\ket{b,\no{e}} \\
& & \qquad + \;
   0.00000099\ket{\no{b},e} + 0.98999901\ket{\no{b},\no{e}}.
\end{array} \]

\noindent The conditional probability table for alarm translates in a
straightforward manner into a channel $c$ from $\{b,\no{b}\} \times
\{e,\no{e}\}$ to $\{a,\no{a}\}$, namely as:
\[ \begin{array}{rclcrcl}
c(b,e)
& = &
0.9999\ket{a} + 0.0001\ket{\no{a}}
& \hspace*{2em} &
c(b,\no{e})
& = &
0.99\ket{a} + 0.01\ket{\no{a}}
\\
c(\no{b},e)
& = &
0.99\ket{a} + 0.01\ket{\no{a}}
& &
c(\no{b},\no{e})
& = &
0.0001\ket{a} + 0.9999\ket{\no{a}}.
\end{array} \]

\noindent The following question is asked in~\cite[Example~3.1
  and~3.2]{Barber12}:
\begin{quote}
Imagine that we are 70\% sure we heard the alarm sounding. What is the
probability of a burglary?
\end{quote}

\noindent Again it is not clear if we should interpret this situation
in terms of improvement (Pearl) or correction (Jeffrey). The latter
seems more natural since there is no `surprise' that needs
correction. Nevertheless, \cite{Barber12} uses the former.

For Jeffrey's approach we translate the $70\%$ certainty into a state
$\rho = 0.7\ket{a} + 0.3\ket{\no{a}}$. We can take the Bayesian
inversion of the channel $c$ wrt.\ the product state
$\sigma\otimes\omega$, giving $c^{\dag}_{\sigma\otimes\omega} \colon
\{a,\no{a}\} \chanto \{b,\no{b}\} \times \{e,\no{e}\}$. Jeffreys' rule
$c^{\dag}_{\sigma\otimes\omega} \gg \rho$ thus gives a distribution on
$\{b,\no{b}\} \times \{e,\no{e}\}$. Taking its first marginal yields
the outcome that is computed in~\cite{Barber12}, namely:
\[ 0.693\ket{b} + 0.307\ket{\no{b}}. \]

For Pearl's approach we translate the $70\%$ certainty into a
predicate $p\colon \{a,\no{a}\} \rightarrow [0,1]$ with $p(a) = 0.7$
and $p(\no{a}) = 0.3$. Pearl's rule $(\sigma\otimes\omega)|_{c \ll p}$
also yields a a distribution on $\{b,\no{b}\} \times \{e,\no{e}\}$,
whose first marginal is:
\[ 0.0229\ket{b} + 0.9771\ket{\no{b}}. \]

\noindent This outcome is obtained in~\cite{JacobsZ19}. It differs
considerably from the previous one --- $69\%$ versus $2\%$ --- and
demonstrates that it is highly relevant which interpretation ---
Jeffreys' or Pearl's --- is chosen.
\end{example}

\begin{example}
\label{ex:Dietrich}
We look at one more illustration, from~\cite{DietrichLB16}, where we
see an interesting combination of Jeffrey's and Pearl's rule. The
setting is: Ann must decide about hiring Bob, whose characteristics
are described in terms of competence ($c$ or $\no{c}$) and experience
($e$ or $\no{e}$). The prior is a joint distribution on the product
space $\{c,\no{c}\}\times\{e,\no{e}\}$ given as:
\[ \begin{array}{rcl}
\omega
& = &
\frac{4}{10}\ket{c,e} + \frac{1}{10}\ket{c,\no{e}} + 
   \frac{1}{10}\ket{\no{c},e} + \frac{4}{10}\ket{\no{c},\no{e}}.
\end{array} \]

\noindent The first marginal of $\omega$ is the uniform distribution
$\frac{1}{2}\ket{c} + \frac{1}{2}\ket{\no{c}}$. It is the base rate
for Bob's competence.

We use the two projection functions $\{c,\no{c}\}
\stackrel{\pi_1}{\longleftarrow} \{c,\no{c}\}\times\{e,\no{e}\}
\stackrel{\pi_2}{\longrightarrow} \{e,\no{e}\}$ as deterministic
channels $\hat{\pi}_{1}$ and $\hat{\pi}_{2}$.

When Ann would learn that Bob has relevant work experience, given by
point evidence $\indic{e}$, her strategy is to factor this in via
Pearl's rule / backward inference: this gives $\omega|_{\hat{\pi}_{2}
  \ll \indic{e}}$, whose first marginal is $\frac{4}{5}\ket{c} +
\frac{1}{5}\ket{\no{c}}$. It is then more likely that Bob is
competent.

Ann reads Bob's letter to find out if he actually has relevant
experience. We quote from~\cite{DietrichLB16}:
\begin{quote}
Bob's answer reveals right from the beginning that his written English
is poor. Ann notices this even before figuring out what Bob says about
his work experience. In response to this unforeseen learnt input, Ann
lowers her probability that Bob is competent from $\frac{1}{2}$ to
$\frac{1}{8}$.  It is natural to model this as an instance of Jeffrey
revision.
\end{quote}

\noindent Bob's poor English is a new state of affairs --- a surprise
--- which translates to a competence state $\rho = \frac{1}{8}\ket{c}
+ \frac{7}{8}\ket{\no{c}}$. This is not something that Ann wants to
\emph{factor in}; no, she wants to \emph{adjust} to this new
situation, so she uses Jeffrey's rule, giving a new joint
state:
\[ \begin{array}{rcccl}
\omega'
& = &
(\hat{\pi}_{2})^{\dag}_{\omega} \gg \rho
& = &
\frac{1}{10}\ket{c,e} + \frac{1}{40}\ket{c,\no{e}} + 
   \frac{7}{40}\ket{\no{c},e} + \frac{7}{10}\ket{\no{c},\no{e}}.
\end{array} \]

\noindent If the letter now tells that Bob has work experience, Ann
will factor this in, in this new situation $\omega'$, giving
$\frac{4}{11}\ket{c} + \frac{7}{11}\ket{\no{c}}$ as first marginal of
$\omega'|_{\hat{\pi}_{1} \ll \indic{e}}$. The likelihood of Bob being
competent is now lower than in the prior state. This example
reconstructs the illustration from~\cite{DietrichLB16} in
channel-based form, with the associated formulations of Pearl's and
Jeffrey's rules, and produces exactly the same outcomes as
in \textit{loc.\ cit.}
\end{example}

\subsection{All things, or nothing else, considered}\label{subsec:specmethod}

We now take a closer look at~\cite{ChanD05,Darwiche09} where the
Jeffrey/Pearl distinction has been described in terms of the way that
soft evidence is described.
\begin{enumerate}
\item In this context, Jeffrey's rule is called ``all things
  considered''~\cite{GoldszmidtP96}; briefly,
  following~\cite[\S3.6.1]{Darwiche09}: ``One method for describing
  soft evidence on event $\beta$ is by stating the new belief in
  $\beta$ after the evidence has been accomodated.''

We make this more concrete in terms of a probability distribution
$\omega\in\Dst(X)$ which is somehow updated to a distribution
$\omega'\in\Dst(X)$. There is an event $E\subseteq X$ whose
``strength'' is given as its validity $q\in [0,1]$ in the updated
state, that is $q =\, \omega'\models\indic{E}$. This validity in the
updated state is thus the way in which softness is specified. This
makes sense from Jeffrey's perspective, since it involves
adjustment/correction.  It is a rather indirect, \textit{post hoc} way
of specifying, but it can be done like this.

We elaborate this situation in the current framework, using the
partition-based special case of Lemma~\ref{lem:deterministic}. The
event $E\subseteq X$ forms a two-element partition of $X$, consisting
of $E$ and its complement $\neg E$, so we take as index set $I =
\{1,2\}$ with function $f\colon X \rightarrow I$ given by $f(x) = 1$
if $x\in E$ and $f(x) = 2$ if $x\not\in E$. The validity $q$ can be
understood as a `state of affairs' distribution $\sigma = q\ket{1} +
(1-q)\ket{2}$ on the index set $I = \{1,2\}$. Then, following the
formula in~\eqref{eqn:deterministic} for Jeffrey's updating with a
partition, we get convex combination:
\[ \begin{array}{rcccl}
\omega'
& = &
\hat{f}^{\,\dag}_{\omega} \gg \sigma
& = &
q\cdot \omega|_{\indic{E}} + (1-q)\cdot \omega|_{\indic{\neg E}}.
\end{array} \]

\noindent By elaborating the definition of
conditioning~\eqref{eqn:conditioning} we get:
\[ \begin{array}{rcl}
\omega'(x)
& = &
\left\{\begin{array}{ll}
\displaystyle\frac{q\cdot \omega(x)}{\omega\models\indic{E}} 
   & \mbox{if }x\in E
\\[+1.5em]
\displaystyle\frac{(1-q)\cdot \omega(x)}{\omega\models\indic{\neg E}} \quad
   & \mbox{if }x\in \neg E
\end{array}\right.
\end{array} \]

\noindent This is precisely Eqn.~(3.20) in~\cite{Darwiche09}.

\smallskip

\item Pearl's rule is called ``nothing else considered''. The strength
  is now given by a ``Bayes factor'' $k > 0$.  Skipping many details,
  we can turn this factor $k$ into a predicate $p\colon I \rightarrow
  [0,1]$ on the index set $I = \{1,2\}$, with $p(1) = r$ and $p(2) =
  \nicefrac{r}{k}$. The number $r$ is some scaling factor that ensures
  that $p$'s values are in the unit interval $[0,1]$. It drops out in
  updating, see
  Lemma~\ref{lem:conditioning}~\eqref{lem:conditioning:scalar}.

We elaborate the technicalities of Pearl's approach, using the above
partition $E, \neg E$ of $X$ over $I = \{1,2\}$. Then we compute
Pearl's update $\omega|_{\hat{f} \ll p}$ step-by-step:
\[ \begin{array}{rcl}
\big(\hat{f} \ll p\big)(x)
& = &
p(f(x))
\hspace*{\arraycolsep}=\hspace*{\arraycolsep}
\left\{\begin{array}{ll}
r \quad  & \mbox{if }x\in E
\\
\nicefrac{r}{k} & \mbox{if }x\not\in E
\end{array}\right.
\\[+1.2em]
\omega \models \hat{f} \ll p
& = &
\sum_{x\in E}\, r\cdot\omega(x) + \sum_{x\not\in E}\, \nicefrac{r}{k}\cdot\omega(x)
\\
& = &
r\cdot(\omega\models \indic{E}) + \nicefrac{r}{k}\cdot (\omega\models \indic{\neg E})
\\[+0.3em]
\omega|_{\hat{f} \ll p}
& = &
\displaystyle\sum_{x\in E}\, \frac{r\cdot(\omega\models \indic{E})}
  {r\cdot(\omega\models \indic{E}) + \nicefrac{r}{k}\cdot (\omega\models \indic{\neg E})}\bigket{x}
\\
& & \qquad +\;\displaystyle
\sum_{x\not\in E}\, \frac{\nicefrac{r}{k}\cdot (\omega\models \indic{\neg E})}
  {r\cdot(\omega\models \indic{E}) + \nicefrac{r}{k}\cdot (\omega\models \indic{\neg E})}\bigket{x}.
\end{array} \]

\noindent We can rewrite the latter formal convex sum as probability
mass function:
\[ \begin{array}{rcl}
\big(\omega|_{\hat{f} \ll p}\big)(x)
& = &
\left\{\begin{array}{ll}
\displaystyle\frac{k\cdot \omega(x)}
   {k\cdot (\omega\models\indic{E}) + (\omega\models\indic{\neg E})} 
   & \mbox{if }x\in E
\\[+1.5em]
\displaystyle\frac{\omega(x)}
   {k\cdot (\omega\models\indic{E}) + (\omega\models\indic{\neg E})} \quad
   & \mbox{if }x\in \neg E
\end{array}\right.
\end{array} \]

\noindent This is Eqn.~(3.25) in~\cite{Darwiche09}.

\auxproof{
Example~\ref{ex:Pearlalarm} as described above occurs
in~\cite[\S3.6.2]{Darwiche09} with Bayes factor $4$. This corresponds
to the predicate $p\colon A \rightarrow [0,1]$ with $p(a) = 0.8$ and
$p(\no{a}) = 0.2$ that we have used in Example~\ref{ex:Pearlalarm}
with Bayes factor ratio $\nicefrac{p(a)}{p(\no{a})} = 4$. The
underlying set $A\times B$ may be partitioned here as $\{a\}\times B$
and $\{\no{a}\}\times B$.

We can consider this example with two-element index set $I = \{1,2\}$,
where $(a,b),(a,\no{b})$ are mapped to 1, and
$(\no{a},b),(\no,\no{b})$ to 2. For hyper conditioning of 
\[ \begin{array}{rcl}
\omega
& = &
0.000095\ket{a,b} + 0.009999\ket{a,\no{b}} + 0.000005\ket{\no{a},b} +
   0.989901\ket{\no{a},\no{b}}
\end{array} \]

\noindent write $r_{1} = 0.000095 + 0.009999$ and $r_{2} = 0.000005 +
0.989901$. Then:
\[ \begin{array}{rcl}
\mathcal{H}(\omega)
& = &
r_{1}\bigket{\nicefrac{0.000095}{r_1}\ket{a,b} + 
   \nicefrac{0.009999}{r_1}\ket{a,\no{b}}} 
\\
& & \qquad +\;
r_{2}\bigket{\nicefrac{0.000005}{r_2}\ket{\no{a},b} +  
   \nicefrac{0.989901}{r_2}\ket{\no{a},\no{b}}}
\end{array} \]

\noindent Pearl's updating with $p$ yields:
\[ \begin{array}{rcl}
\lefteqn{\frac{0.8r_{1}}{0.8r_{1} + 0.2r_{2}}\bigket{\ldots} +
   \frac{0.2r_{1}}{0.8r_{1} + 0.2r_{2}}\bigket{\ldots}}
\\
& \mapsto &
\frac{0.8 \cdot 0.000095}{0.8r_{1} + 0.2r_{2}}\ket{a,b} + 
   \frac{0.8 \cdot 0.009999}{0.8r_{1} + 0.2r_{2}}\ket{a,\no{b}}
\\
& & \qquad +\;
\frac{0.2\cdot 0.000005}{0.8r_{1} + 0.2r_{2}}\ket{\no{a},b} +  
   \frac{0.2 \cdot 0.989901}{0.8r_{1} + 0.2r_{2}}\ket{\no{a},\no{b}}
\\
& = &
0.00036883105790453487\ket{a,b} + 0.038820439452499404\ket{a,\no{b}}
\\
& & \qquad +\;
4.853040235585985e-06\ket{\no{a},b} +  
   0.9608058764493604\ket{\no{a},\no{b}}
\end{array} \]

\noindent This corresponds to the updated distribution in
Example~\ref{ex:Pearlalarm}.
}
\end{enumerate}

\noindent We conclude that, even though the approaches ``all things
considered'' and ``nothing else considered'' take a completely
different route to specifying softness, they still fit in the current
general setting.

\section{Concluding remarks}\label{sec:conclusion}

This paper uses hard maths for soft evidence. It provides a systematic
account of two different forms of probabilistic updating with soft
evidence, namely Jeffrey's rule and Pearl's method. These two
approaches are provided with informal conceptualisations, like:
adjusting to, correction (Jeffrey style), and: factoring in,
improvement (Pearl style). The paper's technical contribution lies in
providing a mathematically precise formulation of Jeffrey's and
Pearl's updating, systematically using the concept of channel. This
makes it possible to reformulate several results from the literature,
notably from~\cite{ChanD05,Darwiche09}, to add new results, and to
describe various (confusing) examples from a uniform perspective.

In the end we briefly suggest a connection between the channel-based
formalism and the cognitive explanation of perception
in~\cite{Hohwy13}. We formalise it as follows: a consistent, relevant
portion of the human mind may be represented by a probability
distribution $\sigma$, forming the internal state at hand. We use a
channel $c$ to translate this internal state into predictions $c \gg
\sigma$ about the outside world. The confrontation of this prediction
with observation leads to an update of the internal state $\sigma$.
In the setting of this paper, the update may happen using Jeffrey's
approach, when $\sigma$ is adjusted/corrected to $c^{\dag}_{\sigma}
\gg \rho$ for an observed external state $\rho$. It may also happen
according to Pearl, so that $\sigma$ is improved to $\sigma|_{c \ll
  p}$ for external evidence $p$ that is factored in.  It remains an
intriguing open question, far beyond the scope of this paper, if this
Jeffrey/Pearl distinction between correcting and improving makes
cognitive sense.

Finally, a question that might arise is whether Jeffrey's/Pearl's
updating can also be described (and distinguished) in continuous
probability. The answer is yes. Pearl's updating is essentially
conditioning and can be done with continuous probability, see
\textit{e.g.}~\cite{Jacobs18c}. Jeffrey's approach involves
disintegration (or Bayesian inversion), which is a rather subtle topic
in a continuous setting, see~\cite{ClercDDG17} (and the references
there) for more information: daggers of channels may not exist, or may
not be determined uniquely (up to null-sets).

\subsection*{Acknowledgements} Thanks to the anonymous reviewers
for their constructive feedback.

\vskip 0.2in

\end{document}